%% file: acl2023.tex
\pdfoutput=1

\documentclass[11pt]{article}

\usepackage[final]{ACL2023}

\usepackage{times}
\usepackage{latexsym}

\usepackage[T1]{fontenc}

\usepackage[utf8]{inputenc}

\usepackage{microtype}

\usepackage{inconsolata}

\usepackage{caption}
\usepackage{bbold}
\usepackage{multirow}
\usepackage{array, makecell} %
\usepackage{dsfont}
\usepackage{listings}
\usepackage{lipsum}
\usepackage{graphicx}
\usepackage{booktabs}
\input{init}
\input{dfn}

\title{\textsc{MultiInstruct}: Improving Multi-Modal Zero-Shot Learning via Instruction Tuning}

\author{Zhiyang Xu$^*$, \quad Ying Shen\thanks{\quad Zhiyang Xu and Ying Shen contributed equally to this work.}, \quad Lifu Huang \\
  Computer Science Department \\
  Virginia Tech \\
  \texttt{\{zhiyangx, yings, lifuh\}@vt.edu} 
  }

\begin{document}
\maketitle

\input{sections/0_abstract.tex}
\input{sections/1_intro.tex}
\input{sections/2_related.tex}
\input{sections/3_method.tex}
\input{sections/4_experimental_setup.tex}
\input{sections/5_results.tex}
\input{sections/6_conclusion.tex}
\input{sections/7_limitations.tex}
\input{sections/8_acknowledgement}
\bibliography{anthology,custom}
\bibliographystyle{acl_natbib}

\appendix
\input{sections/A1_tasks.tex}

\input{sections/A2_attention.tex}

\end{document}

%% file: init.tex
\usepackage{graphicx}               
\usepackage{tabularx}               
\usepackage{booktabs}
\usepackage{amsmath}
\usepackage{stmaryrd}
\usepackage{xcolor}
\usepackage{soul}

\newcolumntype{C}{>{\centering\arraybackslash}X}
\usepackage{multirow}               
\usepackage{diagbox}                
\usepackage{hhline}                 
\usepackage{color,colortbl}         
\usepackage{amsmath}                
\usepackage{amssymb}                
\usepackage{mathtools}              
\usepackage{enumitem}               

\usepackage{subcaption}
\usepackage{caption}
\usepackage{booktabs}
\usepackage{extarrows}
\usepackage{makecell}
\usepackage{cleveref}
\usepackage{hyperref}

 \usepackage{soul}

\definecolor{RoseQuartzBg}{HTML}{F7CAC9}
\definecolor{RoseQuartz}{HTML}{F5A798}
\definecolor{Serenity}{HTML}{92A8D1}
\definecolor{OrangeRed}{rgb}{1.0, 0.27, 0.0}
\definecolor{Red}{rgb}{1.0, 0.0, 0.0}
\definecolor{Turquoise}{HTML}{0F4C81}
\definecolor{Gray}{gray}{0.9}
\usepackage{xparse}
\usepackage{footmisc}

\NewDocumentCommand{\ying}{ mO{} }{\textcolor{teal}{\textsuperscript{\textit{Ying}}\textsf{\textbf{\small[#1]}}}}

\NewDocumentCommand{\lifu}{ mO{} }{\textcolor{blue}{\textsuperscript{\textit{Lifu}}\textsf{\textbf{\small[#1]}}}}

\NewDocumentCommand{\zhiyang}{ mO{} }{\textcolor{Turquoise}{\textsuperscript{\textit{zhiyang}}\textsf{\textbf{\small[#1]}}}}



%% file: dfn.tex
\newcommand*\numtasks{62}
\newcommand*\numcategories{10}

%% file: sections/0_abstract.tex
\begin{abstract}

Instruction tuning, a new learning paradigm that fine-tunes pre-trained language models on tasks specified through instructions, has shown promising zero-shot performance on various natural language processing tasks. However, it has yet to be explored for vision and multimodal tasks. In this work, we introduce \textsc{MultiInstruct}, the first multimodal instruction tuning benchmark dataset that consists of \numtasks{} diverse multimodal tasks in a unified seq-to-seq format covering \numcategories{} broad categories. The tasks are derived from 21 existing open-source datasets and each task is equipped with 5 expert-written instructions. We take OFA~\cite{wang2022unifying} as the base pre-trained model for multimodal instruction tuning, and to further improve its zero-shot performance, we explore multiple transfer learning strategies to leverage the large-scale \textsc{Natural Instructions} dataset~\cite{Swaroop2022naturalinstructions}. Experimental results demonstrate
strong zero-shot performance on various unseen multimodal tasks and the benefit of transfer learning from a text-only instruction dataset.
We also design a new evaluation metric -- \textit{Sensitivity}, to evaluate how sensitive the model is to the variety 
of instructions. Our results indicate that fine-tuning the model on a diverse set of tasks and instructions leads to a reduced sensitivity to variations in instructions for each task\footnote{The dataset, source code, and model checkpoints are publicly available at \url{https://github.com/VT-NLP/MultiInstruct}.}.
\end{abstract}

%% file: sections/1_intro.tex
\section{Introduction}


With the advances in large-scale pre-trained language models (PLMs), recent studies have explored various efficient learning paradigms~\cite{brown2020language,liu2021pre,wei2021flan,xie2021explain-in-context} to generalize PLMs to new tasks without task-specific tuning. Among these, instruction tuning~\cite{wei2021flan} has achieved significant success in zero-shot learning on natural language processing tasks. By fine-tuning a PLM on tasks described through instructions, instruction tuning allows the model to learn to understand and follow the instructions to perform predictions on unseen tasks. Recent advancement in multimodal pre-training~\cite{wang2022unifying, alayrac2022flamingo, bao2022beit, wang2022beit3} has shown the potential of jointly interpreting text and images in a shared semantic space, which further leads us to ask: can the instruction tuning be leveraged to improve the generalizability of Vision-Language pre-trained models on multi-modal and vision tasks?

In this work, we propose \textsc{MultiInstruct}, the first benchmark dataset for multimodal instruction tuning with \numtasks{} diverse tasks from \numcategories{} broad categories, 
including Visual Question Answering~\cite{goyal2017making,suhr2017corpus}, Commonsense Reasoning~\cite{zellers2019recognition,xie2019visual}, 
Visual Relationship Understanding~\cite{krishna2017visual} and so on. We equipped each task with 5 instructions that are written by two experts in natural language processing. As shown in Figure~\ref{fig:teaser}, we formulate all the tasks into a unified sequence-to-sequence format in which the input text, images, instructions, and bounding boxes are represented in the same token space. 

\input{figures/teaser.tex}

We use OFA~\cite{wang2022unifying}\footnote{We use OFA as it was the largest and most powerful open-source multimodal pre-trained model available at the time of our research while other stronger models didn't have publicly available checkpoints at that time.}, a unified model that is pre-trained on a diverse set of multimodal and unimodal tasks in a single Transformer-based sequence-to-sequence framework, as the base pre-trained multimodal language model, and fine-tune it on \textsc{MultiInstruct}. To utilize \textsc{Natural Instructions}~\cite{Swaroop2022naturalinstructions}, a large-scale text-only instruction tuning dataset, we further explore two transfer learning strategies, including \textit{Mixed Instruction Tuning} and \textit{Sequential Instruction Tuning}.
Experimental results demonstrate strong zero-shot performance on various unseen multimodal tasks with instruction tuning and the potential of further improving it by leveraging large-scale text-only instruction datasets. 

As suggested by previous studies~\cite{pavlick2022promptwording, Liu2022fewshot}, PLMs are highly sensitive toward the wording and length of instructions. Thus, we propose a new metric 
-- \textit{Sensitivity}, which measures how sensitive the model is toward the variety of instructions for the same task. Experimental results demonstrate that (1) instruction tuning significantly reduces the sensitivity of OFA to the varying wording of instructions. The more tuning tasks and instructions for each task are introduced, the lower sensitivity tends to be achieved, and (2) transferring from a larger text-only instruction dataset can also significantly reduces the sensitivity of OFA. 

%% file: figures/teaser.tex
\begin{figure*}[t!]
    \centering
\includegraphics[width=\textwidth]{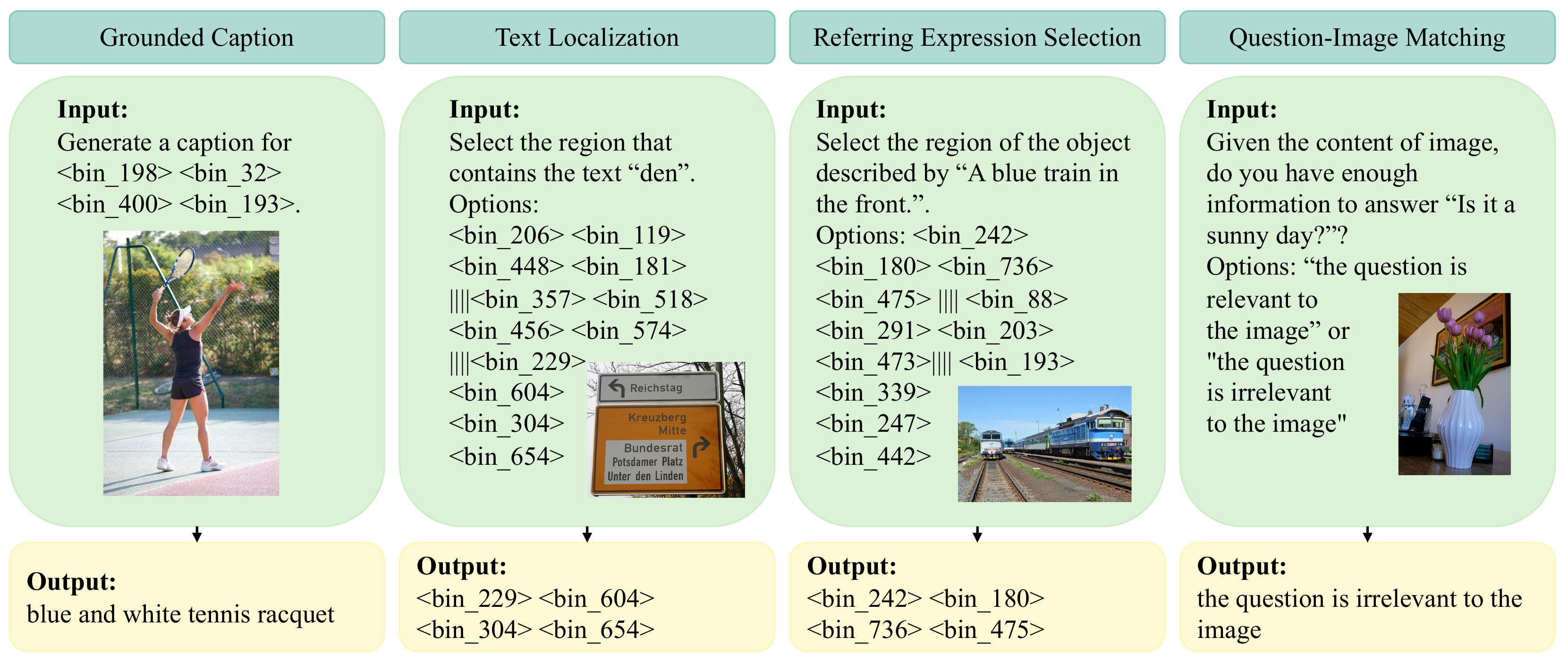}
    \caption{\textbf{Example Instances from \textsc{MultiInstruct} for Four Tasks.} }
    \label{fig:teaser}
\end{figure*}

%% file: sections/2_related.tex
\section{Related Work}
\paragraph{Multimodal Pretraining} Multimodal pretraining \cite{tan2019lxmert,cho2021unifying,singh2022flava,alayrac2022flamingo,wang2022unifying, li2022lavender, li2022UniPerceiverV2} has significantly advanced the vision-language tasks. 
Several recent studies~\cite{cho2021unifying,wang2022unifying, wang2022beit3, lu2022unifiedIO} also started to build a unified pre-training framework to handle a diverse set of cross-modal and unimodal tasks. Among them, VL-T5~\cite{cho2021unifying} tackles vision-and-language tasks with a unified text-generation objective conditioned on multimodal inputs, while OFA~\cite{wang2022unifying} further extends it to image generation tasks by using a unified vocabulary for all text and visual tokens. 
BEIT-3~\cite{wang2022beit3} utilizes a novel shared Multiway Transformer network with a shared self-attention module to align different modalities and provide deep fusion. 
Building on the success of multimodal pretraining, our work focuses on improving the generalization and zero-shot performance on various unseen multimodal tasks through instruction tuning.






\paragraph{Efficient Language Model Tuning} To improve the generalizability and adaptivity of large-scale pre-trained language models, various efficient language model tuning strategies have been proposed recently. Prompt tuning~\cite{liu2021pre,Li2021prefixtuning,han2022ptr,wang2022art,Sanh2022T0} aims to learn a task-specific prompt by reformulating the downstream tasks to the format that the model was initially trained on and has shown competitive performance across various natural language processing applications.
As a special form of prompt tuning, in-context learning~\cite{xie2021explain-in-context,min2021metaicl} takes one or a few examples as the prompt to demonstrate the task. 
Instruction tuning~\cite{wei2021flan} is another simple yet effective strategy to improve the generalizability of large language models. 
\textsc{Natural Instructions}~\cite{Swaroop2022naturalinstructions} is a meta-dataset containing diverse tasks with human-authored definitions, things to avoid, and demonstrations. It has shown effectiveness in improving the generalizability of language models even when the size is relatively small (e.g., BART\_base)~\cite{Swaroop2022naturalinstructions,Wang2022superNaturalInstruct}. 
InstructDial~\cite{prakhar2022instructdial} applies instruction tuning to the dialogue domain and shows significant zero-shot performance on unseen dialogue tasks. While these studies have been successful in text-only domains, it has not yet been extensively explored for vision or multimodal tasks. 

%% file: sections/3_method.tex
\section{\textsc{MultiInstruct}}
\label{sec:multi_instruct}


\subsection{Multimodal Task and Data Collection}

The \textsc{MultiInstruct} dataset is designed to cover a wide range of multimodal tasks that require reasoning among regions, images, and text. These tasks are meant to teach machine learning models to perform various tasks such as object recognition, visual relationship understanding, text-image grounding, and so on by following instructions so that they can perform zero-shot prediction on unseen tasks. 
To build \textsc{MultiInstruct}, we first collect 34 tasks from the existing studies in visual and multimodal learning, covering Visual Question Answering~\cite{goyal2017making,krishna2017visual,zhu2016visual7w,hudson2019gqa,singh2019towards,marino2019ok}, Commonsense Reasoning~\cite{suhr2017corpus,liu2022visual,zellers2019recognition,xie2019visual}, Region Understanding~\cite{krishna2017visual}, Image Understanding~\cite{kafle2017analysis, chiu2020assessing},
Grounded Generation~\cite{krishna2017visual,yu2016modeling,lin2014microsoft}, Image-Text Matching~\cite{lin2014microsoft,goyal2017making}, Grounded Matching~\cite{krishna2017visual,veit2016coco,yu2016modeling}, Visual Relationship~\cite{krishna2017visual,pham2021learning}, 
 Temporal Ordering tasks that are created from WikiHow\footnote{\url{https://www.wikihow.com}.}, and Miscellaneous~\cite{yao2022end,kiela2020hateful,das2017visual,lin2014microsoft,veit2016coco,alam2022medic}. Each of the 34 tasks can be found with one or multiple open-source datasets, which are incorporated into \textsc{MultiInstruct}. Details of each task and their corresponding datasets are shown in ~\Cref{tab:mm_instruct_train1,tab:mm_instruct_train2,tab:mm_instruct_test} in Appendix.


\input{figures/multi_instruct_tasks.tex}


For each of these tasks, we further examine the possibility of deriving new tasks based on the input and output of the original task to augment the task repository. 
For example, \textit{Visual Grounding} requires the model to generate a caption for a given region in the image. We derive two additional tasks from it: \textit{Grounded Caption Selection}, which is a simpler task that requires the model to select the corresponding caption from multiple candidates for the given region, and \textit{Visual Grounding Selection}, which requires the model to select the corresponding region from the provided candidate regions based on a given caption. Compared with \textit{Visual Grounding}, these two new tasks require different skills based on distinct input and output information. 
In this way, we further derived 28 new tasks from the 34 existing tasks. We divide all 62 tasks into 10 broad categories as shown in Figure~\ref{fig:multi_instructions}.



For the existing tasks, we use their available open-source datasets to create instances (i.e., input and output pairs) while for each new task, we create its instances by extracting the necessary information from instances of existing tasks or reformulating them. Each new task is created with 5,000 to 5M instances. We split the 62 tasks into training and evaluation based on the following criteria:
(1) we take the tasks that are similar to the pre-training tasks of OFA~\cite{wang2022unifying} for training; and (2) we select the challenging multimodal tasks that do not overlap with the training tasks for evaluation. 
Table~\ref{tab:modality_stat} and Table~\ref{tab:detailed_stat} in Appendix~\ref{appendix:tasks} show the detailed statistics for the training and evaluation tasks in \textsc{MultiInstruct} and
\Cref{tab:mm_instruct_train1,tab:mm_instruct_train2,tab:mm_instruct_test} show their corresponding datasets.

\subsection{Task Instruction Creation}

We first provide a definition for ``\textit{instruction}'' used in \textsc{MultiInstruct}. An \textit{instruction} is defined with a template that describes how the task should be performed and contains an arbitrary number of placeholders, including \texttt{<TEXT>}, \texttt{<REGION>} and \texttt{<OPTION>}, for the input information from the original task. For example, in the instruction of the Grounded Captioning task, ``Generate a caption for \texttt{<REGION>}'', \texttt{<REGION>} is the placeholder for region-specific information. Note that the placeholder \texttt{<OPTION>} is only used in classification tasks and for some tasks, the input may also include an image that is not included in the instruction and will be fed as a separate input to the model. Figure \ref{fig:teaser} provides several instruction examples for the tasks included in \textsc{MultiInstruct}.

To produce high-quality instructions that accurately convey the intended tasks, we employ an iterative annotation process involving two expert annotators who have a thorough understanding of the task and the dataset.

\noindent\textbf{Step 1:} each annotator first writes 2-3 instructions for each task by giving them 
the specific goals of this task, the format of input data, and 10 example instances randomly sampled from the dataset. The information about the dataset is obtained from the dataset's README file or the publication that introduced the dataset. For newly derived tasks, we provide annotators with task descriptions along with 10 constructed example instances.

\noindent\textbf{Step 2:} to guarantee the quality of the instructions and that they effectively convey the intended tasks, we have each annotator review the instructions created by their peers, checking if they can clearly understand and identify the intended task by just reading the instruction. If any issues are identified, the reviewing annotator provides suggestions and works with the original annotator to revise the instructions.

\noindent\textbf{Step 3:} to ensure the consistency and avoid conflicts or repetition among instructions from different annotators, we have both annotators review the sets of instructions together, identifying any discrepancies or inconsistencies. If any are found, the annotators collaborate to resolve them and create a final set of instructions that accurately and clearly describe the task. In this way, each task will be created with 5 high-quality instructions. 

\noindent\textbf{Step 4:} we repeat steps 1-3 to create 5 instructions for each of the training and evaluation tasks. Finally, both annotators review each task and its instructions and filter out the task that is not representative or overlaps with other tasks.

\subsection{Multimodal Instruction Formatting}
\label{sec:scheme}


To unify the processing of various input/output data types, we follow the method from OFA~\cite{wang2022unifying}, which involves representing images, text, and bounding box coordinates as tokens in a unified vocabulary. Specifically, we apply byte-pair encoding (BPE) \cite{sennrich2016neural} to encode the text input. For the target image, we apply VQ-GAN \cite{esser2021taming} to generate discrete image tokens through image quantization. To represent regions or bounding boxes of an image, we discretize the four corner coordinates into location tokens such as "<bin\_242> <bin\_180> <bin\_736> <bin\_475>" where each location token "<bin\_NUM>" represents a quantized coordinate obtained by dividing the image into 1,000 bins. This approach allows us to convert different types of input into a unified vocabulary.

All tasks in \textsc{MultiInstruct} can then be formulated as natural language sequence-to-sequence generation problems, where the input includes: (1) an image (if there is no input image, a black picture is used as the input); 
and (2) an instruction where the placeholders such as \texttt{<TEXT>}, \texttt{<REGION>} or \texttt{<OPTION>} are filled with specific information of each input instance. Notably, for the \texttt{<OPTION>} of the instructions for classification tasks, we introduce two special tokens for this field: ``\texttt{[Options]}'' to mark the beginning of the option field and ``\texttt{||||}'' to delimit the given options. We concatenate all the options with ``\texttt{||||}'' in the option field and the model will directly generate one option from them. Figure \ref{fig:teaser} provides several examples of the formulated input and illustrates how the original data input is combined with the instruction in the \textsc{MultiInstruct}.

\section{Problem Setup and Models}

\subsection{Problem Setup}

We follow the same instruction tuning setting as the previous study~\cite{wei2021flan} and mainly evaluate the zero-shot learning capabilities of the fine-tuned large language models. Specifically, given a pre-trained multimodal language model $M$, we aim to finetune it on a collection of instruction tasks $T$. 
Each task $t \in T$ is associated with a number of training instances 
$\mathcal D^t=\{(I^t, x^t_j, y^t_j) \in \mathcal{I}^t \times \mathcal X^t \times \mathcal Y^t\}_{j=1}^N$,
where $x^t_j$ denotes the input text, image, region, and options if provided, $y^t_j$ denotes the output of each instance, and $I^t$ represents the set of five task instructions written by experts. The input information from $x^t_j$ will be used to fill in the placeholders in the instruction.

We use OFA~\cite{wang2022unifying} as the pre-trained multimodal model due to its unified architecture and flexible input-output modalities. We finetune it on our \textsc{MultiInstruct} dataset to demonstrate the effectiveness of instruction tuning. Specifically, we use the transformer-based encoder of OFA to encode the instruction along with all necessary information and an optional image, and predict the output with the transformer-based decoder. Given that the training dataset contains many tasks, we mix all the training instances from these tasks and randomly shuffle them. For each instance, we also randomly sample an instruction template for each batch-based training. Note that, 
though some of the training tasks in \textsc{MultiInstruct} are similar to the pre-training tasks of OFA\footnote{Table \ref{tab:ofa_dataset} in Appendix lists the multimodal tasks and dataset used in OFA pre-training. }, we ensure that the evaluation tasks in \textsc{MultiInstruct} do not overlap with either the pre-training tasks in OFA nor the training tasks in \textsc{MultiInstruct}.

\subsection{Transfer Learning from \textsc{Natural Instructions}}
\label{subsec:transfer}
We notice that the scale of \textsc{Natural Instructions}~\cite{Swaroop2022naturalinstructions} is significantly larger than \textsc{MultiInstruct}, indicating the potential of transferring the instruction learning capability from the larger set of natural language tasks to multimodal tasks. We take 832 English tasks in \textsc{Natural Instructions} and explore several simple transfer-learning strategies:
\paragraph{Mixed Instruction Tuning (OFA$_{\text{MixedInstruct}}$)} We combine the instances of  \textsc{Natural Instructions} and \textsc{MultiInstruct} and randomly shuffle them before finetuning OFA with instructions. Note that, each task in \textsc{Natural Instructions} is just associated with one instruction while for each instance from \textsc{MultiInstruct}, we always randomly sample one instruction from the five instructions for each instance of training.

\paragraph{Sequential Instruction Tuning (OFA$_{\text{SeqInstruct}}$)} Inspired by the Pre-Finetuning approach discussed in~\newcite{Armen2021prefinetuning}, we propose a two-stage sequential instruction tuning strategy where we first fine-tune OFA on the \textsc{Natural Instructions} dataset to encourage the model to follow instructions to perform language-only tasks, and then further fine-tune it on \textsc{MultiInstruct} to adapt the instruction learning capability to multimodal tasks. To maximize the effectiveness of the \textsc{Natural Instructions} dataset, we use all instances in English-language tasks to tune the model in the first training stage.

%% file: figures/multi_instruct_tasks.tex
\begin{figure*}[ht!]
    \centering
\includegraphics[width=\textwidth]{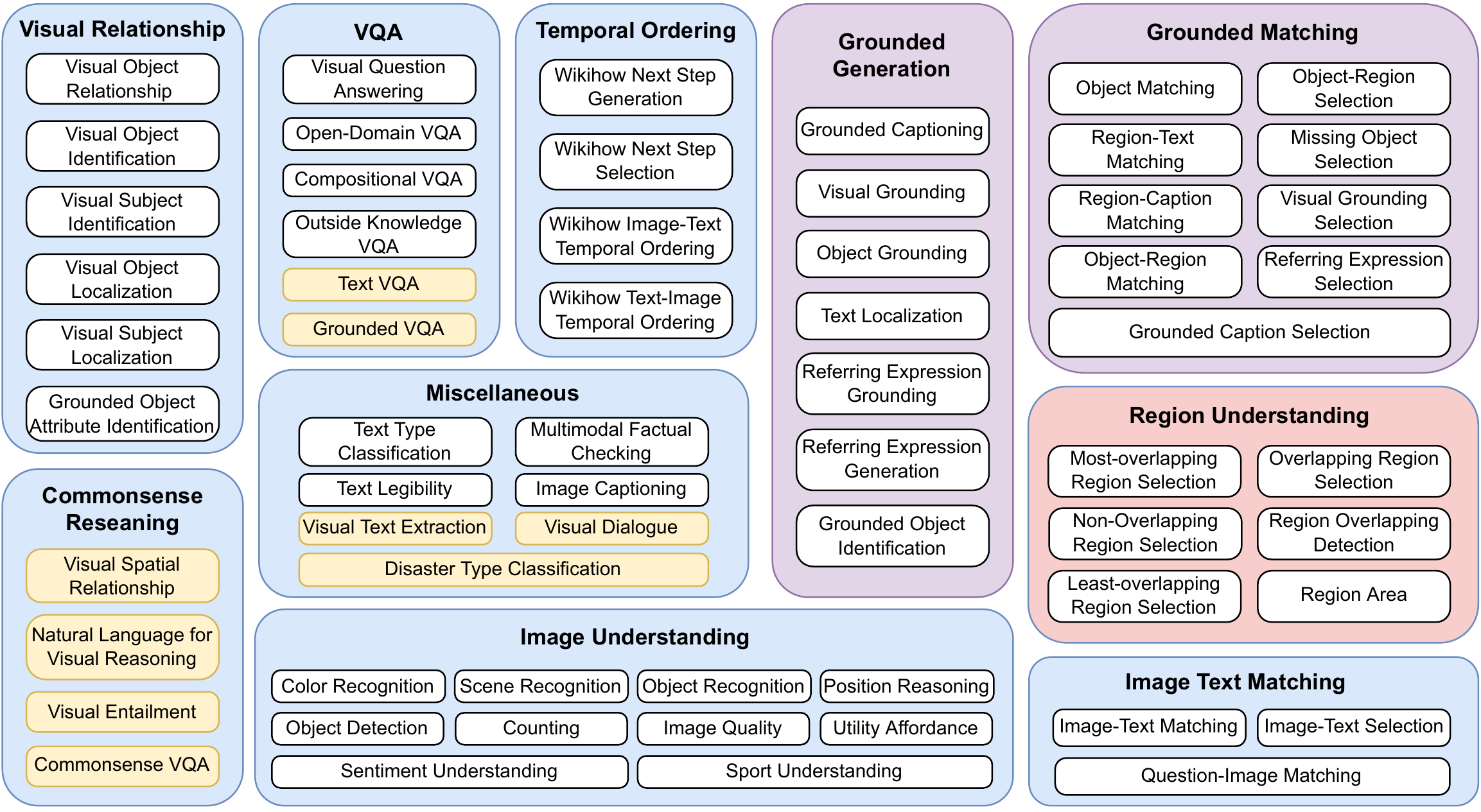}
    \caption{\textbf{Task Groups Included in \textsc{MultiInstruct}.} The yellow boxes represent tasks used for evaluation, while the white boxes indicate tasks used for training.}
    \label{fig:multi_instructions}
\end{figure*}

%% file: sections/4_experimental_setup.tex
\section{Experimental Setup}

\paragraph{Evaluation Metrics}
We report the accuracy for classification tasks and ROUGE-L~\cite{lin2004rouge} for all generation tasks. For the region classification task, we compute the Intersection over Union (IoU) between the generated region and all regions in the options, select the option with the highest IoU as the prediction, and compute accuracy based on this prediction. If the predicted region has no intersection with any of the regions in the options, we treat this prediction as incorrect. 
For classification tasks where the answer is not a single-word binary classification, we also report  ROUGE-L scores following \citet{Swaroop2022naturalinstructions}, which treats all tasks as text generation problems. 
For each task, we conduct five experiments by evaluating the model using one of the five instructions in each experiment. We report the mean and maximum performance and the standard deviation of the performance across all five experiments. We also compute the \textit{aggregated performance} for each model based on the mean of the model's performance on all multimodal and NLP unseen tasks. We use Rouge-L as the evaluation metric for most tasks and accuracy for tasks that only have accuracy as a metric.

In addition, as instruction tuning mainly relies on the instructions to guide the model to perform prediction on various unseen multimodal tasks, we further propose to evaluate how sensitive the model is to the variety of human-written instructions in the same task, which has not been discussed in previous instruction tuning studies but is necessary to understand the effectiveness of instruction tuning. We thus further design a new metric as follows:

\vspace{1mm}

\noindent\textbf{\textit{Sensitivity}} refers to the model's capability of consistently producing the same results, regardless of slight variations in the wording of instructions, as long as the intended task remains the same. Specifically, for each task $t\in T$, given its associated instances  
with  task instructions: $\mathcal D^t=\{(I^t, x^t_j, y^t_j) \in \mathcal{I}^t \times \mathcal X^t \times \mathcal Y^t\}_{j=1}^N$, we formally define \textit{sensitivity} as:
\begin{equation*}
    \mathbb{E}_{t\in T}\left[\frac{\sigma_{i\in I^t}\left[\mathbb{E}_{(x,y)\in \mathcal D^t}[\mathcal L(f_\theta(i,x),y)]\right]}{\mu_{i\in I^t}\left[\mathbb{E}_{(x,y)\in \mathcal D^t}[\mathcal L(f_\theta(i,x),y)]\right]}\right]
\end{equation*}
where $\mathcal{L}$ denotes the evaluation metric such as accuracy or ROUGE-L, $f_\theta(\cdot)$ represents the multimodal instruction-tuned model. 
The standard deviation and mean of the model's performance across all instructions are denoted by $\sigma_{i \in I^t}[\cdot]$ and $\mu_{i \in I^t}[\cdot]$, respectively.


\paragraph{Evaluation datasets}

We evaluate the models on nine unseen multimodal tasks: Text VQA~\cite{singh2019towards}, Grounded VQA~\cite{zhu2016visual7w}, Commonsense VQA~\cite{zellers2019recognition}, Visual Entailment~\cite{xie2019visual}, Visual Spatial Reasoning~\cite{liu2022visual}, Natural Language for Visual Reasoning (NLVR)~\cite{suhr2017corpus}, Visual Text Extraction~\cite{kiela2020hateful}, Visual Dialogue~\cite{das2017visual}, and Disaster Type Classification~\cite{alam2022medic}. 
These tasks belong to three task groups: Commonsense Reasoning, VQA, and Miscellaneous as shown in Figure \ref{fig:multi_instructions}.
Tasks in the Commonsense Reasoning group have no overlap with any training task groups. Tasks in Miscellaneous do not share similarities with other tasks in the group. Although Text VQA and Grounded VQA belong to the VQA task group, they require additional skills such as extracting text from images or generating regions, making them fundamentally different from other tasks in VQA.
In addition to multimodal tasks, we also evaluate the model on 20 NLP tasks collected from the test split of \textsc{Natural Instructions}.



\paragraph{Approaches for Comparison}
We denote the OFA finetuned on \textsc{MultiInstruct} as \textbf{OFA$_{\text{MultiInstruct}}$}, and compare it with the original pre-trained \textbf{OFA}\footnote{\url{https://ofa-beijing.oss-cn-beijing.aliyuncs.com/checkpoints/ofa\_large.pt}}, \textbf{OFA$_{\text{TaskName}}$} which is fine-tuned on \textsc{MultiInstruct} but uses the task name instead of instruction to guide the model to make predictions, and several approaches that leverage the large-scale \textsc{Natural Instructions} dataset, including \textbf{OFA$_{\text{NaturalInstruct}}$} which only fine-tunes OFA on \textsc{Natural Instructions} with instruction tuning, \textbf{OFA$_{\text{MixedInstruct}}$} and \textbf{OFA$_{\text{SeqInstruct}}$} that are specified in Section~\ref{subsec:transfer}. 

More details regarding the evaluation datasets, baseline approaches and training details can be found in Appendix~\ref{append:exp_setup}.

%% file: sections/5_results.tex
\section{Results and Discussion}

\input{tables/zs_vcr.tex}
\input{tables/zs_vte.tex}

\subsection{Effectiveness of Instruction Tuning on \textsc{MultiInstruct}}
\label{sec:instruction_effect}


We evaluate the zero-shot performance of various approaches on all the unseen evaluation tasks, as shown in Table~\ref{table:zs_vcr} and \ref{table:zs_vte}. Our results indicate that OFA$_\text{MultiInstruct}$ significantly improves the model's zero-short performance over the original pre-trained OFA model across all unseen tasks and metrics, demonstrating the effectiveness of multimodal instruction tuning on \textsc{MultiInstruct}. As seen in Table \ref{table:zs_vte}, OFA achieves extremely low (nearly zero) zero-shot performance on the Grounded VQA task, which requires the model to generate region-specific tokens in order to answer the question. By examining the generated results, we find that OFA, without instruction tuning, failed to follow the instruction and produce results that contain region tokens. However, by fine-tuning OFA on \textsc{MultiInstruct}, the model is able to better interpret and follow the instructions to properly generate the expected output.
Additionally, OFA$_\text{MultiInstruct}$ outperforms OFA$_\text{TaskName}$ on all unseen tasks, particularly on the Grounded VQA task, where OFA$_\text{TaskName}$ achieves nearly zero performance. This suggests that the performance gain of OFA$_\text{MultiInstruct}$ mainly comes from instructions rather than multi-task training.

\subsection{Impact of Transfer Learning from \textsc{Natural Instructions}}
One key question in multimodal instruction tuning is how to effectively leverage the large-scale text-only \textsc{Natural Instructions} dataset to enhance the zero-shot performance on multimodal tasks. We observe that only fine-tuning OFA on \textsc{Natural Instructions} actually degrades the model's zero-shot performance on almost all multimodal tasks, as shown by comparing OFA$_\text{NaturalInstruct}$ and OFA in Table~\ref{table:zs_vcr} and \ref{table:zs_vte}. One potential reason for this decline in performance is that during fine-tuning on the text-only dataset, the model learns to focus more on text tokens and attend less to image tokens. To verify this assumption, we compare the attention of text tokens on image tokens between OFA$_\text{NaturalInstruct}$ and other methods and observe that text tokens attend much less to image tokens after fine-tuning on the \textsc{Natural Instructions} dataset. The detailed explanations and analysis can be found in Appendix~\ref{attention_analysis}.


Another observation is that although our transfer learning methods do not lead to significant performance gains over OFA$_\text{MixedInstruct}$, both OFA$_\text{SeqInstruct}$ and OFA$_\text{MixedInstruct}$ achieve lower standard deviation on 6 out of 9 unseen multimodal tasks compared with OFA$_\text{MultiInstruct}$, demonstrating the potential benefits of the much larger text-only instruction datasets to multimodal instruction tuning.

\subsection{Impact of Increasing Multimodal Instruction Task Clusters}
\label{subsec:num_tasks}

\input{figures/num_task_result.tex}

To evaluate the impact of the number of tasks clusters for instruction tuning, we start with the task groups shown in Figure \ref{fig:multi_instructions} and group them into five larger clusters: (1) \texttt{Img Und} (VQA +  Image Understanding), (2) \texttt{Grounding} (Grounded Matching + Grounded Generation), (3) \texttt{MISC, ITM} (Temporal Ordering + Miscellaneous + Image Text Matching), (4) \texttt{Relation} (Visual Relationship), (5) \texttt{Region} (Region Understanding), together with (6) \texttt{NLP}, a collection of NLP tasks from \textsc{Natural Instructions}. We measure the change in both the aggregated performance and \textit{sensitivity} of OFA$_\text{MixedInstruct}$ as we gradually add the task clusters for training. 

As we increase the number of task clusters, we observe an improvement in both the mean and maximum aggregated performance and a decrease in \textit{sensitivity}, as shown in Figure \ref{fig:num_task_result}. Note that low \textit{sensitivity} indicates that the model can produce consistent results despite variations in the wording of instructions. These results suggest that increasing the number of task clusters improves the model's performance on unseen tasks and leads to more consistent outputs. The results also support the effectiveness of our proposed \textsc{MultiInstruct} dataset.

\subsection{Effect of Diverse Instructions on Instruction Tuning}
\label{subsec:num_instruct}
We hypothesize that using a diverse set of instructions for each task during multimodal instruction tuning can improve the model's zero-shot performance on unseen tasks and reduce its \textit{sensitivity} to variation in the instructions. To test this hypothesis, we train an OFA model on \textsc{MultiInstruct} with a single fixed instruction template per task and compare its performance with OFA finetuned on 5 different instructions. 
As shown in Table \ref{tab:num_instruct_result}, OFA finetuned on 5 instructions achieves much higher aggregated performance on all evaluation tasks and shows lower \textit{sensitivity}. These results demonstrate the effectiveness of increasing the diversity of instructions and suggest that future work could explore crowd-sourcing or automatic generation strategies to create even more diverse instructions for instruction tuning.



\input{tables/num_instruct_result.tex}


\subsection{Effect of Fine-tuning Strategies on Model \textit{Sensitivity}}

In Section~\ref{subsec:num_tasks} and~\ref{subsec:num_instruct}, we have shown that the more tasks and instructions used for instruction tuning, the lower \textit{sensitivity} the model will achieve toward the variations in instructions for each task. We further investigate the impact of fine-tuning and transfer learning strategies on model sensitivity. Figure~\ref{fig:robuestness} shows the averaged \textit{sensitivity} of each model across all multimodal unseen tasks. The original OFA exhibits significantly higher sensitivity to variations in instructions compared to models fine-tuned on instruction datasets, indicating that multimodal instruction tuning significantly improves the model's capability on interpreting instructions, even with varying wordings. In addition, by transferring the large-scale \textsc{Natural Instructions} dataset to \textsc{MultiInstruct}, \textit{sensitivity} is also reduced by a large margin, highlighting the benefit of fine-tuning the model on a larger instruction dataset, regardless of different formats and modalities.

\input{figures/robustness.tex}

\section{Zero-Shot Performance on NLP Tasks}

So far, our focus has been on evaluating the zero-shot performance of multimodal tasks. In this section, we investigate the effect of multimodal instruction tuning on the performance of text-only tasks. To do this, we evaluate all our approaches on 20 natural language processing (NLP) tasks from the default test split in \textsc{Natural Instructions}\footnote{\url{https://github.com/allenai/natural-instructions}}. The detailed task list can be found in Appendix~\ref{sec:nlp_test_tasks}.  

\input{tables/zs_nlp.tex}

As shown in Table~\ref{table:zs_nlp}, OFA$_\text{MultiInstruct}$ outperforms OFA, despite the instruction tuning dataset and the unseen dataset are in different modalities. This suggests that multimodal instruction tuning can help improve the zero-shot performance on NLP tasks. In addition, we observe that OFA$_\text{NaturalInstruct}$ achieves the best performance on NLP tasks and OFA$_\text{MixedInstruct}$ is more effective in preserving the zero-shot capability gained from \textsc{Natural Instructions} on NLP tasks compared to OFA$_\text{SeqInstruct}$.
Based on the results in \Cref{table:zs_vcr,table:zs_vte,table:zs_nlp}, we conclude that OFA$_\text{MixedInstruct}$ is able to achieve overall best aggregated performance on all multimodal and NLP tasks and shows much lower \textit{sensitivity} towards variations in the wording of instructions, making it the most promising approach.




%% file: tables/zs_vcr.tex
\begin{table*}[!t]
  \centering
  \resizebox{\linewidth}{!}{%
  \begin{tabular}{l cccc c cc c cc c cc}
  \toprule
   \multirow{3}{*}{\centering \textbf{Model}} & \multicolumn{4}{c}{\textbf{Commonsense VQA}}  && \multicolumn{2}{c}{\textbf{Visual Entailment}} && \multicolumn{2}{c}{\textbf{Visual Spatial Reasoning}} 
   && \multicolumn{2}{c}{\textbf{NLVR}} \\
   \cline{2-5} \cline{7-8} \cline{10-11} \cline{13-14} 
  & \multicolumn{2}{c}{RougeL} & \multicolumn{2}{c}{ACC} && \multicolumn{2}{c}{ACC} && \multicolumn{2}{c}{ACC} && \multicolumn{2}{c}{ACC}\\
  \cline{2-5} \cline{7-8} \cline{10-11} \cline{13-14}
  & Max & Avg $\pm$ \text{Std} & Max & Avg $\pm$ \text{Std} && Max & Avg$\pm$ \text{Std} && Max & Avg$\pm$ \text{Std} && Max & Avg$\pm$ \text{Std} \\
  \midrule
  OFA & 17.93 & 14.97 $\pm$ 4.30 & 0.73 & 0.40 $\pm$0.29 && 49.99 & 41.86  $\pm$ 10.99 && 54.99 & 35.29 $\pm$ 22.21 && 56.06 & 52.10 $\pm$ 3.35  \\
  OFA$_\text{TaskName}$ & 48.99 & - & 29.01 & - && 55.70 & - && 53.76 & - && 55.35 & -  \\
  OFA$_\text{MultiInstruct}$ & \textbf{52.01} & \textbf{50.60} $\pm$ 1.12 & \textbf{33.01} & 31.17 $\pm$ 1.59 && \textbf{55.96} &\textbf{55.06} $\pm$0.76 && \textbf{55.81} &\textbf{53.90} $\pm$1.38 && 56.97 &56.18 $\pm$ 0.95 \\
  \midrule
  \rowcolor{Gray} \multicolumn{14}{l}{\textbf{Transfer Learning from \textsc{Natural Instructions}}} \\
  OFA$_\text{NaturalInstruct}$ & 27.15 & 14.99 $\pm$ 9.12 & 7.35 & 2.04 $\pm$ 3.01 && 33.28 & 14.86 $\pm$ 16.68 && 51.44 & 36.44 $\pm$ 20.72 && 56.06 & 35.98 $\pm$ 21.64   \\
  OFA$_\text{MixedInstruct}$ &  50.40 & 49.34 $\pm$ 1.04 & 31.31 & 30.27 $\pm$ 0.94 && 54.63 & 53.74 $\pm$ 0.97 && 55.13 & 52.61 $\pm$ 1.64 && 56.67 & 55.96 $\pm$ 0.48 \\
  OFA$_\text{SeqInstruct}$ & 50.93 & 50.07 $\pm$ 1.07 & 32.28 & \textbf{31.23} $\pm$ 1.09 && 53.66 & 52.98 $\pm$ 0.56 && 54.86 & 53.11 $\pm$ 1.45 && \textbf{57.58} & \textbf{56.63} $\pm$ 0.66  \\
  \bottomrule
  \end{tabular}}
  \caption{\textbf{Zero-shot Performance on Multimodal Commonsense Reasoning.} The best performance is in \textbf{bold}.}
  \label{table:zs_vcr}
  \end{table*}  

%% file: tables/zs_vte.tex
\begin{table*}[!t]
  \centering
  \resizebox{\linewidth}{!}{%
  \begin{tabular}{l cc c cc c cc c cc c cc}
  \toprule
   \multirow{3}{*}{\centering \textbf{Model}} & \multicolumn{2}{c}{\textbf{Text VQA}} &&\multicolumn{2}{c}{\textbf{Grounded VQA}} && \multicolumn{2}{c}{\textbf{Visual Text Extraction}} && \multicolumn{2}{c}{\textbf{Visual Dialogue}} && \multicolumn{2}{c}{\textbf{Disaster Type Classification}} \\
   \cline{2-3} \cline{5-6} \cline{8-9} \cline{11-12}  \cline{14-15}  
  & \multicolumn{2}{c}{RougeL} && \multicolumn{2}{c}{Acc} && \multicolumn{2}{c}{RougeL} && \multicolumn{2}{c}{RougeL} && \multicolumn{2}{c}{ACC}  \\
  \cline{2-3} \cline{5-6}  \cline{8-9}  \cline{11-12}  \cline{14-15}  
  &  Max & Avg$\pm$ \text{Std} && Max & Avg$\pm$ \text{Std} && Max & Avg$\pm$ \text{Std} &&  Max & Avg $\pm$ \text{Std} && Max & Avg $\pm$ \text{Std}  \\
  \midrule
  OFA & 15.21 & 9.30 $\pm$ 5.42 && 0.02 & 0.00 $\pm$ 0.01 && 36.31 & 17.62 $\pm$ 16.82 && 45.46 & 28.71 $\pm$ 9.81  && 14.30 & 9.64 $\pm$ 4.34 \\
  OFA$_\text{TaskName}$ & 23.80 & - && 0.00 & -&& 36.30 & - && 25.18 & - && 62.65 & -\\
OFA$_\text{MultiInstruct}$ & \textbf{27.22}& 26.46 $\pm$ 0.83 && \textbf{64.32} & 47.22  $\pm$ 23.08 && \textbf{74.35} & \textbf{62.43} $\pm$11.56 && \textbf{46.38}&32.91 $\pm$7.59 && 64.88 &56.00 $\pm$12.96\\
\midrule
\rowcolor{Gray} \multicolumn{15}{l}{\textbf{Transfer Learning from \textsc{Natural Instructions}}} \\
OFA$_\text{NaturalInstruct}$ &5.59 & 5.40 $\pm$ 0.24 && 0.00 & 0.00 $\pm$ 0.00 && 5.65 & 1.24 $\pm$ 2.48 && 30.94 & 27.91 $\pm$ 2.16   && 56.64 & 38.21 $\pm$ 15.35  \\
OFA$_\text{MixedInstruct}$ & 24.15 & 23.67 $\pm$ 0.47 && 63.79 & \textbf{54.99} $\pm$ 18.16 && 62.43 & 46.56 $\pm$ 14.92 && 46.08 & \textbf{38.02} $\pm$ 5.25  && \textbf{68.31} & \textbf{64.31} $\pm$ 2.39 \\
OFA$_\text{SeqInstruct}$ & 27.03 & \textbf{26.67} $\pm$ 0.47 && 64.19 & 54.46 $\pm$ 15.96  && 71.63 & 60.62 $\pm$ 12.31 && 46.17 & 35.10 $\pm$ 6.92  && 64.46 & 57.89 $\pm$ 9.51 \\
  \bottomrule
  \end{tabular}}
  \caption{\textbf{Zero-shot Performance on Question Answering and Miscellaneous.} The best performance is in \textbf{bold}.}
  \label{table:zs_vte}
  \end{table*}

%% file: figures/num_task_result.tex
\begin{figure}[t]
    \centering
\includegraphics[width=0.85\columnwidth]{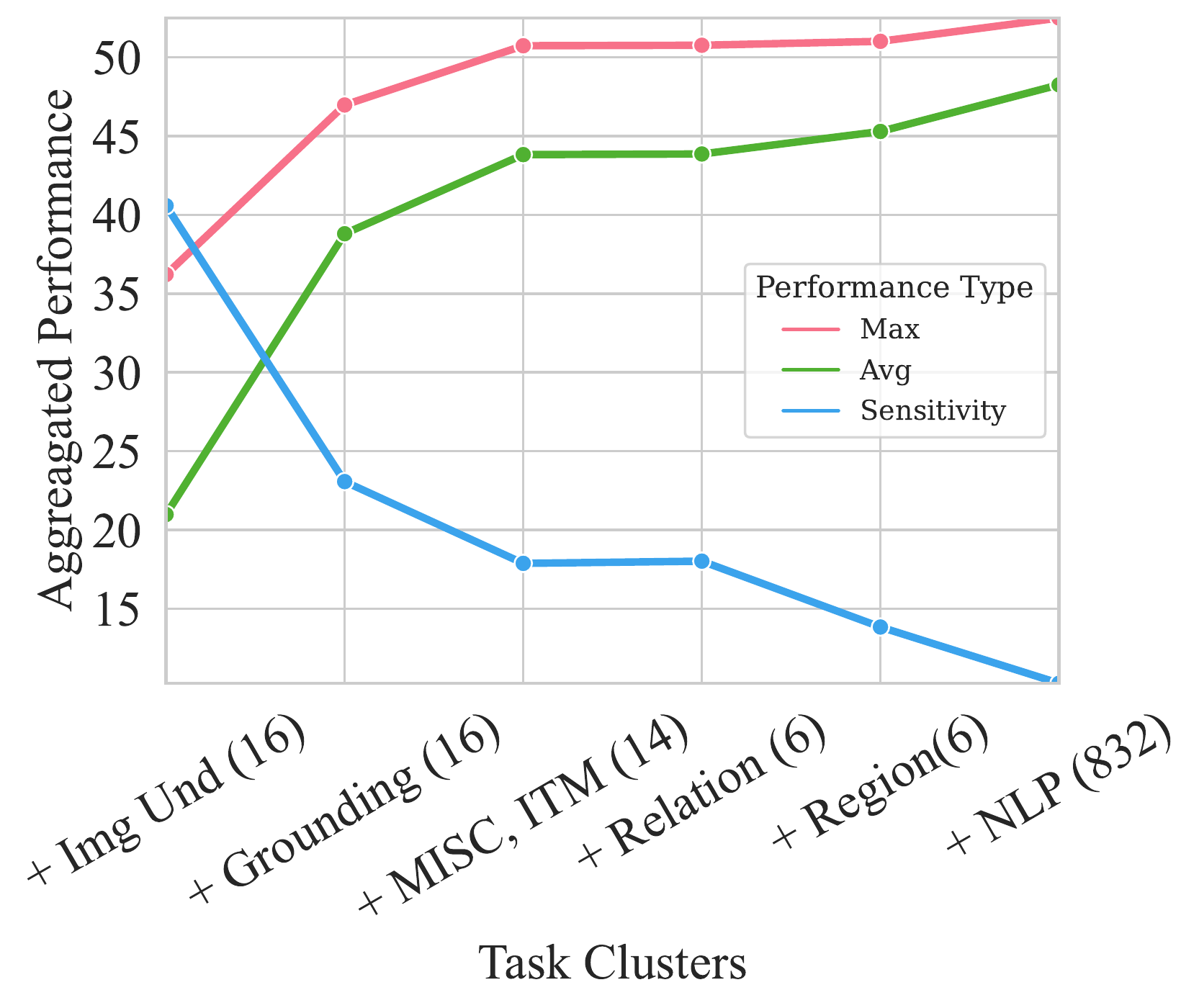}
    \caption{\textbf{Model Performance as the Number of Multimodal Instruction Task Clusters Increases.} The number in the parenthesis of each cluster denotes the number of tasks.}
    \label{fig:num_task_result}
\end{figure}

%% file: tables/num_instruct_result.tex
\begin{table}[!t]
\scriptsize
\begin{center}
\resizebox{\linewidth}{!}{%
\begin{tabular}{l|c|c}
\toprule
\# of Instructions & Aggregated Performance $\uparrow$ & \textit{Sensitivity} $\downarrow$   \\
\midrule
1 Instruction & 42.81 & 24.62 \\
5 Instructions & \textbf{47.82} & \textbf{10.45} \\
\bottomrule
\end{tabular}
}
\caption{\textbf{Effect of Different Number of Instructions.} Performance of OFA$_\text{MultiInstruct}$ finetuned on different numbers of instructions.}
\label{tab:num_instruct_result}
\end{center}
\end{table}

%% file: figures/robustness.tex
\begin{figure}[tb!]
    \centering
\includegraphics[scale=0.3]{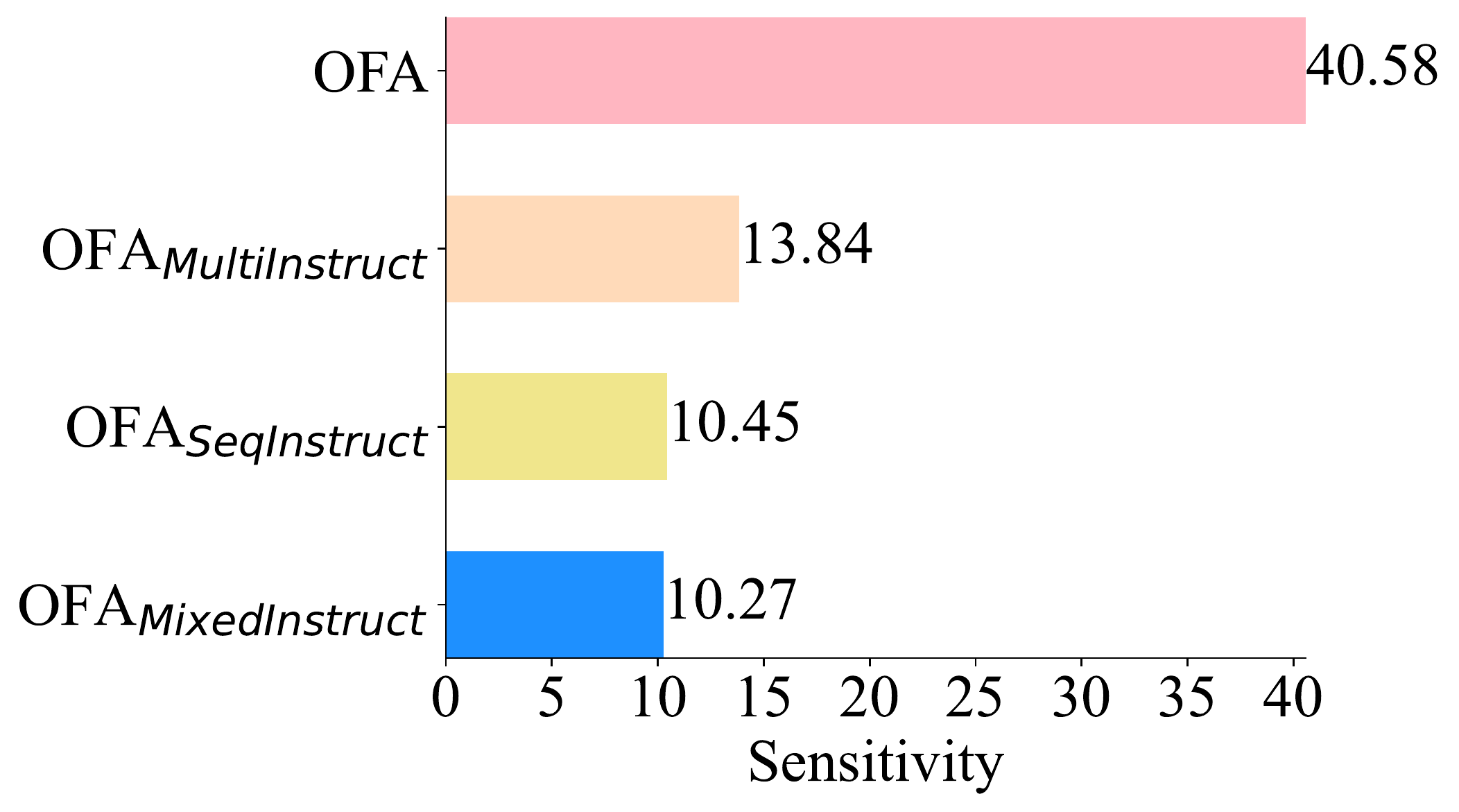}
    \caption{\textbf{Model \textit{Sensitivity} on Unseen Evaluation Tasks.} Lower is better. 
    } 
    \label{fig:robuestness}
\end{figure}

%% file: tables/zs_nlp.tex
\begin{table}[ht]
  \centering
  \resizebox{\linewidth}{!}{%
  \begin{tabular}{l c}
  \toprule
   \textbf{Model} & \textbf{RougeL}   \\
   %
  \midrule
  OFA & 2.25 \\
OFA$_\text{MultiInstruct}$ & 12.18  \\
\midrule
    
\rowcolor{Gray}\multicolumn{2}{l}{\textbf{Transfer Learning from \textsc{Natural Instructions}} \hfill}   \\
    OFA$_\text{NaturalInstruct}$ &  \textbf{43.61} \\
    OFA$_\text{MixedInstruct}$ & 43.32  \\
    OFA$_\text{SeqInstruct}$ & 30.79 \\
  \bottomrule
  \end{tabular}}
  \caption{\textbf{Zero-shot Performance on NLP tasks.} The performance is reported in Rouge-L and the best performance is in \textbf{bold}.}
  \label{table:zs_nlp}
  \end{table}

%% file: sections/6_conclusion.tex
\section{Conclusion}
\vspace{-5pt}
We present a new large-scale multi-modal instruction tuning benchmark dataset -- \textsc{MultiInstruct}, which covers a wide variety of vision and multimodal tasks while each task is associated with multiple expert-written instructions. By finetuning OFA~\cite{wang2022unifying}, a recently state-of-the-art multimodal pre-trained language model, on \textsc{MultiInstruct} with instruction tuning, its zero-shot performance on various unseen multimodal tasks is significantly improved. We also explore several transfer learning techniques to leverage the much larger text-only \textsc{Natural Instructions} dataset and demonstrate its benefit. 
Moreover, we design a new evaluation metric \textit{Sensitivity} to assess the model's sensitivity towards the variations in the wording of instructions. Results show that the model becomes less sensitive to these variations after being fine-tuned on a variety of tasks and instructions.






%% file: sections/7_limitations.tex
\section*{Limitations}
\paragraph{Limitations of Data Collection}
Our proposed dataset only targets English language tasks. Future work should explore multimodal instruction tuning in a more diverse language setting and augment our \textsc{MultiInstruct} with multi-multilingual tasks. 
In addition, our current dataset mainly focuses on vision-language tasks. Datasets from more diverse modalities should be considered such as audio~\cite{Panayotov2015Librispeech,Gemmeke2017Audioset, You2022audioData} and video~\cite{Soomro2012video, Ionescu2014video}. While we have built a novel multimodal instruction dataset containing 62 tasks, the number of tasks and associated instructions remains limited. To address this, future research could consider utilizing crowd-sourcing or automatic generation and augmentation techniques to increase the variety of instructions available. 

\paragraph{Limitations of Experiments and Evaluation}
Our work is the first to explore instruction tuning on multimodal tasks and shows improved performance compared to baseline methods. However, there is still room for improvement, specifically in utilizing text-only instruction datasets. Future research could explore alternative architectures and stronger vision-language pre-trained models, or develop additional training loss functions to better utilize these unimodal instruction datasets.  
Additionally, we only used OFA as the baseline model as it was the largest open-source multimodal pre-trained model available when we conducted this research. As more and stronger multimodal pre-trained models being publicly available, it would be interesting to conduct a thorough comparison between models with different sizes. Finally, we take the first step to define \textit{sensitivity} as a metric to evaluate the robustness of the models on understanding and following human-written instructions, which can be a potential standard metric for all the following instruction-tuning studies. However, it's only based on the variation of model performance across different instructions for the same task. In the future, we will consider more broad factors, e.g., the model's capability to understand different instructions for different tasks (Inter-task sensitivity), to further improve the \textit{sensitivity} metric for instruction tuning.

%% file: sections/8_acknowledgement.tex
\section*{Acknowledgments}
This research is based upon work supported by the U.S. DARPA KMASS Program \# HR001121S0034. The views and conclusions contained herein are those of the authors and should not be interpreted as necessarily representing the official policies, either expressed or implied, of DARPA or the U.S. Government. The U.S. Government is authorized to reproduce and distribute reprints for governmental purposes notwithstanding any copyright annotation therein.

%% file: sections/A1_tasks.tex

\section{Tasks Defined in \textsc{MultiInstruct}}
\label{appendix:tasks}
Table~\ref{tab:modality_stat} shows the distribution of input and output modalities for both training and evaluation tasks in \textsc{MultiInstruct}, and Table~\ref{tab:detailed_stat} shows the detailed statistics for all the training and evaluation tasks separately. ~\Cref{tab:mm_instruct_train1,tab:mm_instruct_train2,tab:mm_instruct_test} provide a comprehensive list of the 62 tasks included in \textsc{MultiInstruct}, along with one example of instruction for each task. 

\input{tables/modality_stat.tex}
\input{tables/detailed_stat.tex}
\input{tables/mm_instructions_train1.tex}

\input{tables/mm_instructions_train2.tex}

\input{tables/mm_instructions_test.tex}

\section{More Details for Experimental Setup}
\label{append:exp_setup}
\subsection{Multimodal Evaluation Datasets}
\label{sec:test_data}

\paragraph{Text VQA \cite{singh2019towards}} requires models to read and reason about the text in an image to answer questions based on them. 

\paragraph{Grounded VQA \cite{zhu2016visual7w}} requires models to answer the questions about an image, with the answers being specific visual regions within the image.

\paragraph{Commonsense VQA \cite{zellers2019recognition}} requires the model to answer a multiple-choice question that requires commonsense reasoning about an image. Both the question and answers are presented in a combination of natural language and references to specific image regions within the image.

\paragraph{Visual Entailment \cite{xie2019visual}} requires the model to determine whether the image semantically entails the text.

\paragraph{Natural Language for Visual Reasoning (NLVR) \cite{suhr2017corpus}} requires the model to answer a question that requires  visual and set-theoretic reasoning on a synthetic image.

\paragraph{Visual Text Extraction} is a new task derived from Hateful Memes \cite{kiela2020hateful} dataset. This task requires the model to extract the text that appears in the image.

\paragraph{Visual Dialogue \cite{das2017visual}} requires the model to answer a question given an image and a dialogue history.

\paragraph{Disaster Type Classification \cite{alam2022medic}} requires the model to determine the disaster type based on the image.

\subsection{NLP Evaluation Tasks}
\label{sec:nlp_test_tasks}
Below are the task names of the 20 NLP tasks that we used to test the zero-shot performance of all the methods. The 20 NLP tasks are from the default test split of the \textsc{Natural Instructions} dataset. During testing, we leverage the 'Definition' of the task as an instruction and prepend it with each input.\\ \\
task1624\_disfl\_qa\_question\_yesno\_classification, task133\_winowhy\_reason\_plausibility\_detection,
task569\_recipe\_nlg\_text\_generation,
task1631\_openpi\_answer\_generation,
task957\_e2e\_nlg\_text\_generation\_generate,
task1386\_anli\_r2\_entailment,
task393\_plausible\_result\_generation,
task670\_ambigqa\_question\_generation,
task890\_gcwd\_classification,
task1534\_daily\_dialog\_question\_classification,
task1388\_cb\_entailment,
task190\_snli\_classification,
task1533\_daily\_dialog\_formal\_classification,
task1598\_nyc\_long\_text\_generation,
task199\_mnli\_classification,
task1439\_doqa\_cooking\_isanswerable,
task1409\_dart\_text\_generation,
task1529\_scitail1.1\_classification,
task648\_answer\_generation,
task050\_multirc\_answerability

\subsection{Approaches for Comparison}
\paragraph{OFA~\cite{wang2022unifying}} denotes the original pre-trained OFA model without any fine-tuning. Here, we use OFA-large\footnote{\url{https://ofa-beijing.oss-cn-beijing.aliyuncs.com/checkpoints/ofa\_large.pt}} which contains 472M parameters and was trained on 8 tasks shown in Table~\ref{tab:ofa_dataset}. As reported in~\newcite{wang2022unifying}, OFA has demonstrated certain zero-shot capability on unseen multimodal tasks.

\paragraph{\textbf{OFA$_{\text{TaskName}}$}} is finetuned on \textsc{MultiInstruct} but it does not use the instructions we created for the tasks. Instead, we prepend the task name to each input and use a semicolon to separate the task name and the input. For a fair comparison, we still keep the two special tokens ``\texttt{[Options]}'' and ``\texttt{||||}'' for the option field.

\paragraph{OFA$_{\text{MultiInstruct}}$}  only fine-tunes OFA on our newly introduced \textsc{MultiInstruct} dataset with instruction tuning.

\paragraph{OFA$_{\text{NaturalInstruct}}$} only fine-tunes OFA on the large-scale \textsc{Natural Instructions} dataset~\cite{Swaroop2022naturalinstructions,Wang2022superNaturalInstruct} with instruction tuning. To ensure a fair comparison, we evaluate this baseline on instruction templates that removed all specific tokens, including ``\texttt{[Options]}'' and ``\texttt{||||}'', since the model being tested has not been exposed to these specific tokens during instruction-tuning. We want to ensure that the evaluation is not biased in favor of models that have seen these tokens during training.

\paragraph{OFA$_{\text{MixedInstruct}}$} fine-tunes OFA on the mix of the large-scale \textsc{Natural Instructions} ~\cite{Swaroop2022naturalinstructions,Wang2022superNaturalInstruct} and \textsc{MultiInstruct} dataset with instruction tuning.

\paragraph{OFA$_{\text{SeqInstruct}}$} sequentially fine-tunes OFA on the large-scale \textsc{Natural Instructions} ~\cite{Swaroop2022naturalinstructions,Wang2022superNaturalInstruct} and \textsc{MultiInstruct} dataset with instruction tuning.

\input{tables/ofa_datasets.tex}

\subsection{Training Details}
We set the maximum length of input tokens to 1024 and the maximum target length to 512. For image preprocessing, we strictly follow the process in the OFA. Please refer to the original paper for more details. We train the models on 8 Nvidia A100 GPUs with a batch size 8 per GPU, a learning rate of 1e-05, and float16 enabled for 3 epochs for all the setups and datasets. We run all the experiments once.

%% file: tables/modality_stat.tex
\begin{table}[ht!]

\begin{center}
\resizebox{\columnwidth}{!}{%
\begin{small}
\begin{tabular}{c | c | c || c | c | c || c | c}
\toprule
\multicolumn{3}{c}{\textbf{Input modality}} & \multicolumn{3}{c}{\textbf{Output Modality}} & \# of Training & \# of Testing \\
Image & Text & Region & Image & Text & Region \\
\midrule
\checkmark & & & &\checkmark & &  1 &0 \\
\checkmark & \checkmark & & &\checkmark & & 14 & 5 \\
\checkmark & &\checkmark & &\checkmark & & 9 &1 \\
\checkmark & &\checkmark & & &\checkmark  & 2 &0 \\
\checkmark & \checkmark & & & &\checkmark & 3 & 1  \\
\checkmark & \checkmark & \checkmark & &\checkmark & & 9 & 0 \\
\checkmark & \checkmark & \checkmark & & &\checkmark & 1 &0 \\
\bottomrule
\end{tabular}
\end{small}
}
\end{center}
\caption{Distribution of input and output modalities for all the tasks in \textsc{MultiInstruct}.}
\label{tab:modality_stat}
\vskip -0.1in
\end{table}

%% file: tables/detailed_stat.tex
\begin{table}[ht!]

\begin{center}
\resizebox{\columnwidth}{!}{%

\begin{tabular}{l | c | c }
\toprule
& Train & Eval \\ 
\midrule
Average \# of Tokens per Instruction &  14.67 & 9.37\\
Averaged \# of Character per Instruction &  85.78 & 58.77\\
Average Levenshtein Distance of Instructions & 63.63 & 54.74\\
\# of Instructions per Task & 5 & 5\\
\# of Classification Tasks & 21 & 3\\
\# of Generation Tasks & 19& 4\\
\# of Existing Tasks &  19& 7 \\
\# of Created Datasets &  21& 0 \\
\bottomrule
\end{tabular}}

\end{center}
\caption{Detailed statistics in \textsc{MultiInstruct}.}
\label{tab:detailed_stat}
\vskip -0.1in
\end{table}

%% file: tables/mm_instructions_train1.tex
\begin{table*}[ht!]
\tiny
\begin{center}
\scriptsize
\begin{tabular}{p{0.08\linewidth} | p{0.1\linewidth} | p{0.14\linewidth} | p{0.45\linewidth} | p{0.03\linewidth}}
\toprule
  Category & Task Name & Dataset & Description & Exist   \\
\midrule
 \multirow{10}{0.1\linewidth}{\hfil VQA} & Open-Domain VQA & VQAv2~\cite{goyal2017making}, Visual Genome~\cite{krishna2017visual} & Answer the question \texttt{<QUESTION>} based on the content of the given image. & \checkmark\\ \cmidrule{2-5}
 & VQA & Visual7w~\cite{zhu2016visual7w} &Answer a visual question \texttt{<QUESTION>} by selecting an answer from given options. \texttt{<OPTION>}& \checkmark\\
\cmidrule{2-5}
 &Compositional VQA & GQA \cite{hudson2019gqa} & Answer a compositional question based on the content of the given image. Question: \texttt{<QUESTION>}& \checkmark\\
\cmidrule{2-5}
 & Outside Knowledge VQA & OK-VQA~\cite{marino2019ok} & Based on your knowledge, \texttt{<QUESTION>}? & \checkmark \\
\midrule
\midrule
\multirow{17}{0.1\linewidth}{\hfil Grounded  Generation} & Grounded Captioning & Visual Genome~\cite{krishna2017visual}& Given the region \texttt{<REGION>} in the image, generate a caption for that region. & \checkmark\\
\cmidrule{2-5}
& Visual Grounding & Visual Genome~\cite{krishna2017visual}& Given a caption \texttt{<TEXT>} for some region in the image, identify the region and generate its bounding box. & \checkmark\\
\cmidrule{2-5}
& Grounded Object Identification & MSCOCO~\cite{lin2014microsoft}& Identify the type of an object in \texttt{<REGION>}. & \checkmark\\
\cmidrule{2-5}
& Object Grounding &MSCOCO~\cite{lin2014microsoft}& What are the regions containing the object \texttt{[TEXT]}? & $\times$ \\
\cmidrule{2-5}
& Referring Expression Grounding &RefCOCO~\cite{yu2016modeling}& Locate a region in an image based on the referring expression \texttt{[TEXT]}. & \checkmark\\ 
\cmidrule{2-5}
& Referring Expression Generation &RefCOCO~\cite{yu2016modeling}& Generate the referring expression for an object in region \texttt{<REGION>}. & \checkmark\\
\cmidrule{2-5}
& Text Localization & COCO-Text~\cite{veit2016coco} & Select a region from options that contain the text \texttt{<TEXT>} in the image. \texttt{<OPTION>} & \checkmark\\
\midrule
\midrule
\multirow{18}{0.1\linewidth}{\centering Region Understanding} & Most-Overlapping Region Selection & Visual Genome~\cite{krishna2017visual} & Given the region \texttt{<REGION>}, decide which region in the options overlaps most with given region. \texttt{<OPTION>} & $\times$ \\ \cmidrule{2-5}
& Non-Overlapping Region Selection & Visual Genome~\cite{krishna2017visual} & Which option does not share common area with \texttt{<REGION>}? \texttt{<OPTION>} & $\times$ \\ \cmidrule{2-5}
& Least-Overlapping Region Selection & Visual Genome~\cite{krishna2017visual} & "Which option has the least shared area with \texttt{<REGION>}?\texttt{<OPTION>}& $\times$ \\ \cmidrule{2-5}
& Overlapping Region Selection & Visual Genome~\cite{krishna2017visual} & Which region from options that has common area with \texttt{<REGION>}? \texttt{<OPTION>}& $\times$ \\ \cmidrule{2-5}
& Region Overlapping Detection & Visual Genome~\cite{krishna2017visual} & Does \texttt{<REGION1>} share common area with \texttt{<REGION2>}? \texttt{<OPTION>} & $\times$ \\ \cmidrule{2-5}
& Region Area & Visual Genome~\cite{krishna2017visual} & Compute the area of \texttt{<REGION>}. & $\times$ \\ 
\midrule
\midrule
\multirow{19}{0.1\linewidth}{\hfil Grounded Matching} & Region-Caption Matching & Visual Genome~\cite{krishna2017visual} & Decide if the caption matches the given region \texttt{<REGION>} in the image. & $\times$ \\ \cmidrule{2-5}
& Grounded Caption Selection &Visual Genome~\cite{krishna2017visual}& Given a region \texttt{<REGION>} in the image, select a caption from given options for that region. \texttt{<OPTION>}& $\times$ \\ \cmidrule{2-5}
& Visual Grounding Selection &Visual Genome~\cite{krishna2017visual}& Given a caption \texttt{<TEXT>} for some region in the image, select the region from the options. \texttt{<OPTION>} & $\times$  \\\cmidrule{2-5}
& Referring Expression Selection & RefCOCO~\cite{yu2016modeling}& Select a region from options based on the referring expression \texttt{<TEXT>}. \texttt{<OPTION>} & $\times$ \\
\cmidrule{2-5}
& Object-Region Matching & MSCOCO~\cite{lin2014microsoft}& Does region \texttt{<REGION>} contain the object \texttt{<TEXT>}? & $\times$\\ \cmidrule{2-5}
& Object-Region Selection &MSCOCO~\cite{lin2014microsoft}& Select the region containing the given object \texttt{<TEXT>}. \texttt{<OPTION>}& $\times$ \\ \cmidrule{2-5}
& Object Matching &MSCOCO~\cite{lin2014microsoft}& Do objects in region \texttt{<REGION1>} and region \texttt{<REGION2>} have the same type? & $\times$ \\ \cmidrule{2-5}
& Missing Object Selection &MSCOCO~\cite{lin2014microsoft}& Select an object from options that does not appear in any of the given regions \texttt{<REGION>}. \texttt{<OPTION>} & $\times$ \\ \cmidrule{2-5}
& Region-Text Matching &COCO-Text~\cite{veit2016coco}& Does  region \texttt{<REGION>} contain the text \texttt{<TEXT>}? & $\times$\\ 

\bottomrule
\end{tabular}
\caption{\textbf{Detailed Group of Training Tasks Included in \textsc{MultiInstruct}.} The complete list of 53 multi-modal tasks, along with examples of the instructions for each task. The existing tasks are indicated with \checkmark, while the newly derived tasks are indicated using $\times$.}
\label{tab:mm_instruct_train1}
\end{center}
\end{table*}

%% file: tables/mm_instructions_train2.tex
\begin{table*}[ht!]
\tiny
\begin{center}
\scriptsize
\begin{tabular}{p{0.08\linewidth} | p{0.1\linewidth} | p{0.14\linewidth} | p{0.45\linewidth} | p{0.03\linewidth}}
\toprule
  Category & Task Name & Dataset & Description & Exist   \\
\midrule
 \multirow{21}{0.1\linewidth}{\centering Image Understanding} & Color Recognition & TDIUC \cite{kafle2017analysis} & Answer the question: \texttt{<QUESTION>} based on the color of an object. \texttt{<OPTION>} &\checkmark \\ \cmidrule{2-5}
& Object Detection &  TDIUC \cite{kafle2017analysis} & This task asks you to identify if an object appears in the image. \texttt{<QUESTION>}\texttt{<OPTION>} &\checkmark\\ \cmidrule{2-5}
& Object Recognition & TDIUC \cite{kafle2017analysis} & In this task you are asked a question about the type of an object in the image. \texttt{<QUESTION>}\texttt{<OPTION>} &\checkmark\\ \cmidrule{2-5}
& Scene Recognition &  TDIUC \cite{kafle2017analysis} & Look at the environment in the image and answer the question accordingly. \texttt{<QUESTION>}\texttt{<OPTION>}&\checkmark\\ \cmidrule{2-5}
& Counting &  TDIUC \cite{kafle2017analysis} & Question: \texttt{<QUESTION>} Please answer the question by counting the object mentioned in the question. \texttt{<OPTION>}&\checkmark\\ \cmidrule{2-5}
& Sentiment Understanding & TDIUC \cite{kafle2017analysis} & Question: \texttt{<QUESTION>}\texttt{<OPTION>} Please answer the question by interpreting the sentiment in the image. &\checkmark\\ \cmidrule{2-5}
& Position Reasoning &  TDIUC \cite{kafle2017analysis} & In this task, you need to analyze the position of objects in an image and answer the following question. \texttt{<QUESTION>}\texttt{<OPTION>}&\checkmark \\ \cmidrule{2-5}
& Utility Affordance &  TDIUC \cite{kafle2017analysis} & Please take a look at the picture and answer the following question by thinking about what each object in the picture can be used for. \texttt{<QUESTION>}\texttt{<OPTION>} &\checkmark\\ \cmidrule{2-5}
& Sport Understanding &  TDIUC \cite{kafle2017analysis} & There are some sports taking place in the image.\texttt{<QUESTION>}\texttt{<OPTION>}& \checkmark  \\ \cmidrule{2-5}
& Image Quality & IQA~\cite{chiu2020assessing}& Select a reason from the options to explain why the image quality is bad. \texttt{<OPTION>} &\checkmark\\ 
\midrule
\midrule
\multirow{13}{0.1\linewidth}{\centering Visual Relationship}  &Object Relationship & Visual Genome~\cite{krishna2017visual} & What is the relationship between the subject in region \texttt{<REGION1>} and object in region \texttt{<REGION2>}? &\checkmark \\ \cmidrule{2-5}
& Visual Object Identification &Visual Genome~\cite{krishna2017visual} &Given the subject in region \texttt{<REGION>}, what is the object that has a relationship \texttt{<TEXT>} with that subject? & $\times$\\ \cmidrule{2-5}
& Visual Subject Identification & Visual Genome~\cite{krishna2017visual} & Given the object in region \texttt{<REGION>}, what is the subject that has a relationship \texttt{<TEXT>} with that object? & $\times$\\ \cmidrule{2-5}
& Visual Object Localization & Visual Genome~\cite{krishna2017visual} & Given the subject in region \texttt{<REGION>}, where is the object in the image that has relationship \texttt{<TEXT>} with the subject? & $\times$\\ \cmidrule{2-5}
& Visual Subject Localization &Visual Genome~\cite{krishna2017visual} &Given the object in region \texttt{<REGION>}, where is the subject in the image that has relationship \texttt{<TEXT>} with the object? & $\times$\\ 
\cmidrule{2-5}
& Grounded Image Attribute Identification & VAW~\cite{pham2021learning}& Decide which option is the attribute of the object in the region \texttt{<REGION>}. \texttt{<OPTION>} &\checkmark \\ 

\midrule
\midrule
\multirow{6}{0.1\linewidth}{\centering Image-Text Matching} &Image-Text Matching &MSCOCO~\cite{lin2014microsoft}& Decide if the text matches the image.& $\times$ \\
\cmidrule{2-5}
& Question-Image Matching &VQAv2~\cite{goyal2017making}& Decide if the image contains an answer to the question \texttt{<QUESTION>}. & $\times$ \\
\cmidrule{2-5}
& Image-Text Selection &MSCOCO~\cite{lin2014microsoft}& Select the text that best matches the image. \texttt{<OPTION>} & $\times$ \\
\midrule
\midrule

\multirow{8}{0.1\linewidth}{Miscellaneous} & Multimodal Factual Checking & MOCHEG~\cite{yao2022end} & Decide if the claim can be supported by the given image and the context. &\checkmark\\ \cmidrule{2-5}
& Text Legibility & COCO-Text~\cite{veit2016coco}& Decide if the text in the given region is legible. &\checkmark\\ \cmidrule{2-5}
& Text Type Classification &COCO-Text~\cite{veit2016coco}& Read the text in the given region and determine the type of text from options. &\checkmark\\ \cmidrule{2-5}
& Image Captioning &MSCOCO~\cite{lin2014microsoft}& Generate a sentence to describe the content of the image. & \checkmark \\
 \midrule
\midrule
\multirow{11}{0.1\linewidth}{Temporal Ordering} 
& Wikihow Next Step Generation & WikiHow \footnote{https://www.wikihow.com \label{wiki}}
& For task \texttt{<TASK>}, given the history steps and the current step with its corresponding image, what is the next step for this task? \newline \texttt{<HISTORY>} & $\times$ \\ \cmidrule{2-5}
& Wikihow Next Step Selection & WikiHow  & For task \texttt{<TASK>}, select the immediate next step to the step specified by the image. & $\times$ \\ \cmidrule{2-5}
& Wikihow Text-Image Temporal Ordering & WikiHow & For the task \texttt{<TASK>}, given the current step \texttt{<STEP>}, decide if the content of the image is the next or previous step. & $\times$ \\ \cmidrule{2-5}
&  Wikihow Image-Text Temporal Ordering & WikiHow & For the task \texttt{<TASK>}, given the current step specified by the image, decide if the step \texttt{<STEP>} is the next or previous step. & $\times$ \\

\bottomrule
\end{tabular}
\caption{\textbf{(Continued) Detailed Group of Training Tasks Included in \textsc{MultiInstruct}.} The complete list of 53 multi-modal tasks, along with examples of the instructions for each task. The existing tasks are indicated with \checkmark, while the newly derived tasks are indicated using $\times$.}
\label{tab:mm_instruct_train2}
\end{center}
\end{table*}

%% file: tables/mm_instructions_test.tex
\begin{table*}[ht!]
\scriptsize
\begin{center}
\begin{tabular}{p{0.08\linewidth} | p{0.1\linewidth} | p{0.14\linewidth} | p{0.45\linewidth} | p{0.03\linewidth}}
\toprule
  Category & Task Name & Dataset & Description & Exist   \\
\midrule
 \multirow{4}{0.1\linewidth}{\hfil VQA} & Text VQA & Text VQA~\cite{singh2019towards} &  There is some text on the image. Answer \texttt{<QUESTION>} based on the text in the image. & \checkmark  \\
\cmidrule{2-5}
 &Grounded VQA & Visual7W \cite{zhu2016visual7w} & Which region is the answer to \texttt{<QUESTION>}? \texttt{<OPTION>}. & \checkmark  \\ 
 
\midrule
\midrule
\multirow{10}{0.1\linewidth}{Commonsense Reasoning} &Natural Language for Visual Reasoning & NLVR~\cite{suhr2017corpus} & Decide if the sentence \texttt{<TEXT>} correctly describes the geometric relationships of objects in a synthesized image. & \checkmark\\
\cmidrule{2-5}
 &Visual Spatial Reasoning & VSR~\cite{liu2022visual}& Decide if the proposed spatial relationship between two objects in an image is "True" or "False" & \checkmark\\ \cmidrule{2-5}
 
& Visual Entailment & SNLI-VE~\cite{xie2019visual} & Can you conclude \texttt{<TEXT>} from the content of image? Select your answer from the options. \texttt{<OPTION>} & \checkmark \\ \cmidrule{2-5}
& Commonsense Visual Question Answering & VCR~\cite{zellers2019recognition} & Look at the image and the regions in the question, \texttt{<QUESTION>}? \texttt{<OPTION>}. & \checkmark \\ 

\midrule
\midrule
\multirow{8}{0.1\linewidth}{\hfil Miscellaneous} & Visual Text Extraction & Hateful Memes~\cite{kiela2020hateful}  & What is the text written on the image? & $\times$ \\
\cmidrule{2-5}
 &Visual Dialogue & Visual Dialogue~\cite{das2017visual} & Given the image and the dialog history below:\newline\texttt{<HISTORY>}\newline \newline\texttt{<QUESTION>}? & \checkmark  \\ \cmidrule{2-5}

 & Disaster Type Classification & MEDIC \cite{alam2022medic}  & What disaster happens in the image? \texttt{<OPTION>} & \checkmark \\
\bottomrule
\end{tabular}
\caption{\textbf{Detailed Group of Evaluation Tasks Included in \textsc{MultiInstruct}.} The complete list of 9 multi-modal tasks, along with examples of the instructions for each task. The existing tasks are indicated with \checkmark, while the newly derived tasks are indicated using $\times$. }
\label{tab:mm_instruct_test}
\end{center}
\end{table*}

%% file: tables/ofa_datasets.tex
\begin{table}[ht!]

\begin{center}
\resizebox{\columnwidth}{!}{%
\begin{small}
\begin{tabular}{l|c}
\toprule
Dataset Name & Task Name   \\
\midrule
Conceptual Caption 12M (CC12M) & Image Captioning  \\
 Conceptual Captions (CC3M) & Image Captioning  \\
 MSCOCO image captions (COCO) & Image Captioning \\
 Visual Genome Captions (VG Captions) & Image Captioning \\
 VQAv2 & Visual Question Answering \\
 VG-QA (~COCO)& Visual Question Answering \\
 GQA (VG) & Visual Question Answering \\
 RefCOCO & Visual Grounding\\
 RefCOCO+ & Visual Grounding \\
 RefCOCOg & Visual Grounding \\
 VG captions & Visual Grounded Captioning\\
 OpenImages & Object Detection \\
 Object365 & Object Detection \\
 VG & Object Detection \\
 COCO & Object Detection \\
 OpenImages & Image Infilling\\
 YFCC100M & Image Infilling\\
 ImageNet-21K & Image Infilling\\
 

\bottomrule
\end{tabular}
\end{small}
}
\end{center}
\caption{\textbf{Multimodal Pre-training Tasks in OFA.}}
\label{tab:ofa_dataset}
\vskip -0.1in
\end{table}

%% file: sections/A2_attention.tex
\section{Attention Analysis} \label{attention_analysis}
\input{figures/A_attention.tex}

In Section \ref{sec:instruction_effect}, we have demonstrated that fine-tuning OFA with \textsc{Natural Instructions} alone results in a decline in its zero-shot performance. In this section, we examine one possible reason for this decline by examining if fine-tuning the model on a text-only instruction dataset causes it to give less attention to image inputs.

To understand this, we conduct an analysis of the self-attention layers within the OFA encoder. The OFA encoder comprises 12 self-attention layers, each with 16 attention heads. We denote the input to self-attention layer $l$ as $h^{(l)} = [x_1^{(l)}, \dots, x_p^{(l)}, \dots x_L^{(l)}]$, where $L$ is the length of sequence. The input $h^{(0)} = [x_1^{(0)}, \dots, x_I^{(0)}, x_{I+1}^{(0)}, \dots x_{I+T}^{(0)}]$ to the first self-attention layer is actually the concatenation of image embeddings and text embeddings, where $I$, $T$ is the length of image and text embeddings respectively. For ease of understanding and simplicity, we have altered the naming conventions and refer to $x_p^{l}, p=[1, ..., I]$ as image states and $x_p^{l}, p=[I+1, ..., I+T]$ as text states.

For each self-attention layer, we first compute the attention given to the image states in relation to text states for each attention head. Specifically, for each text state as the query, we sum its attention scores on image states (i.e. the attention scores where the text state is the query and image states are the keys). We then compute the text-to-image attention across all text states. Finally, we average the text-to-image across all attention heads. This results in a text-to-image attention score for each self-attention layer. 

Figure \ref{fig:attn} illustrates the results of text-to-image attention scores on three unseen multimodal tasks: Text VQA, Visual Entailment, and Visual Text Extraction. The results on all three unseen tasks show that, in all self-attention layers of the OFA encoder, OFA$_\text{NaturalInstruct}$ has significantly lower text-to-image attention scores compared to other models. This decrease is particularly pronounced in the first two self-attention layers. This suggests that fine-tuning the model on a text-only instruction dataset leads to a reduction in the attention paid to image inputs, which may explain the decline in zero-shot performance.

%% file: figures/A_attention.tex
\begin{figure*}[tbh!]
    \begin{subfigure}[b]{0.33\textwidth}
        \includegraphics[width=\textwidth]{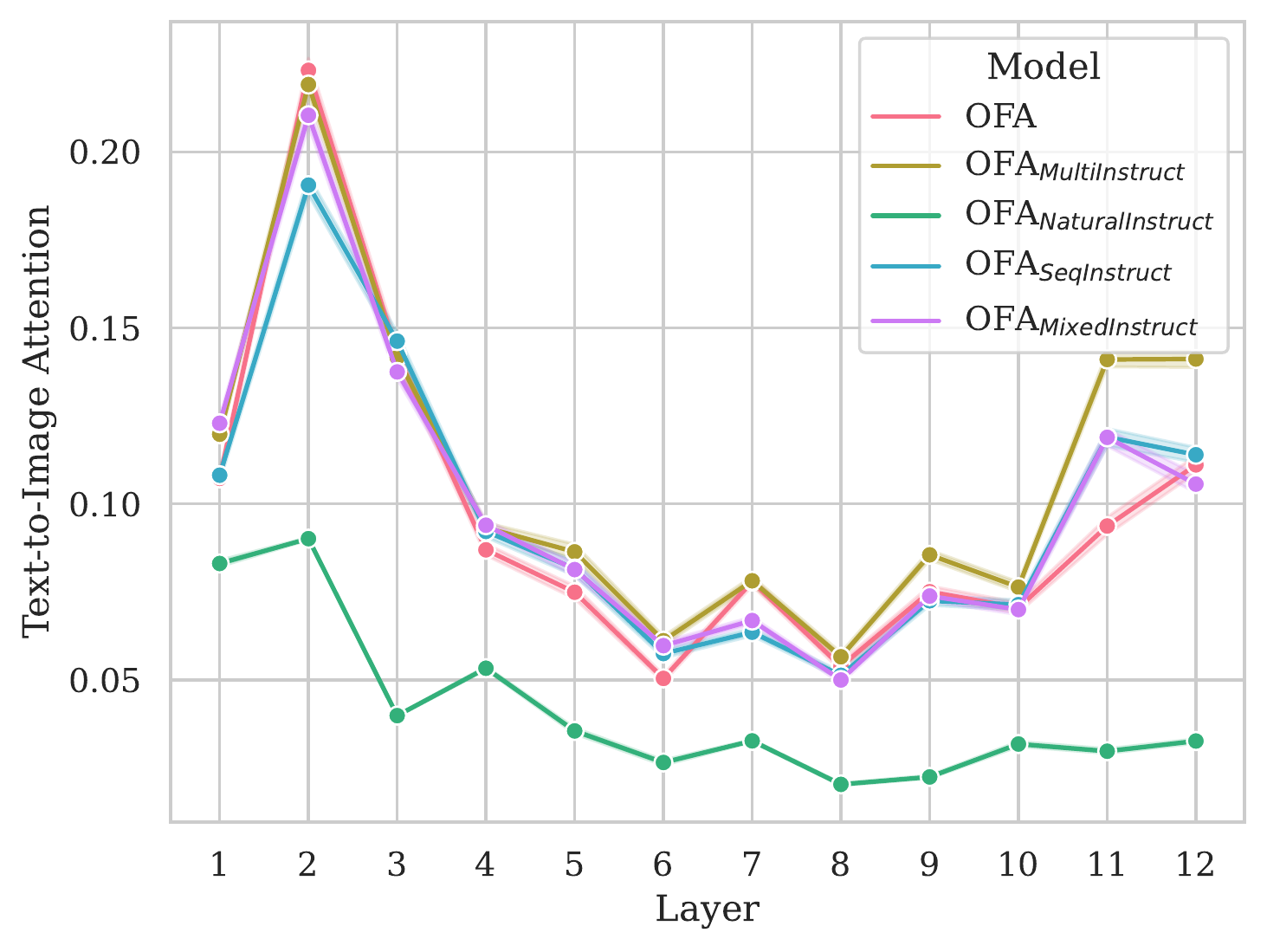}
        \caption{Text VQA}
    \end{subfigure}%
    \hfill
    \begin{subfigure}[b]{0.33\textwidth}
        \includegraphics[width=\textwidth]{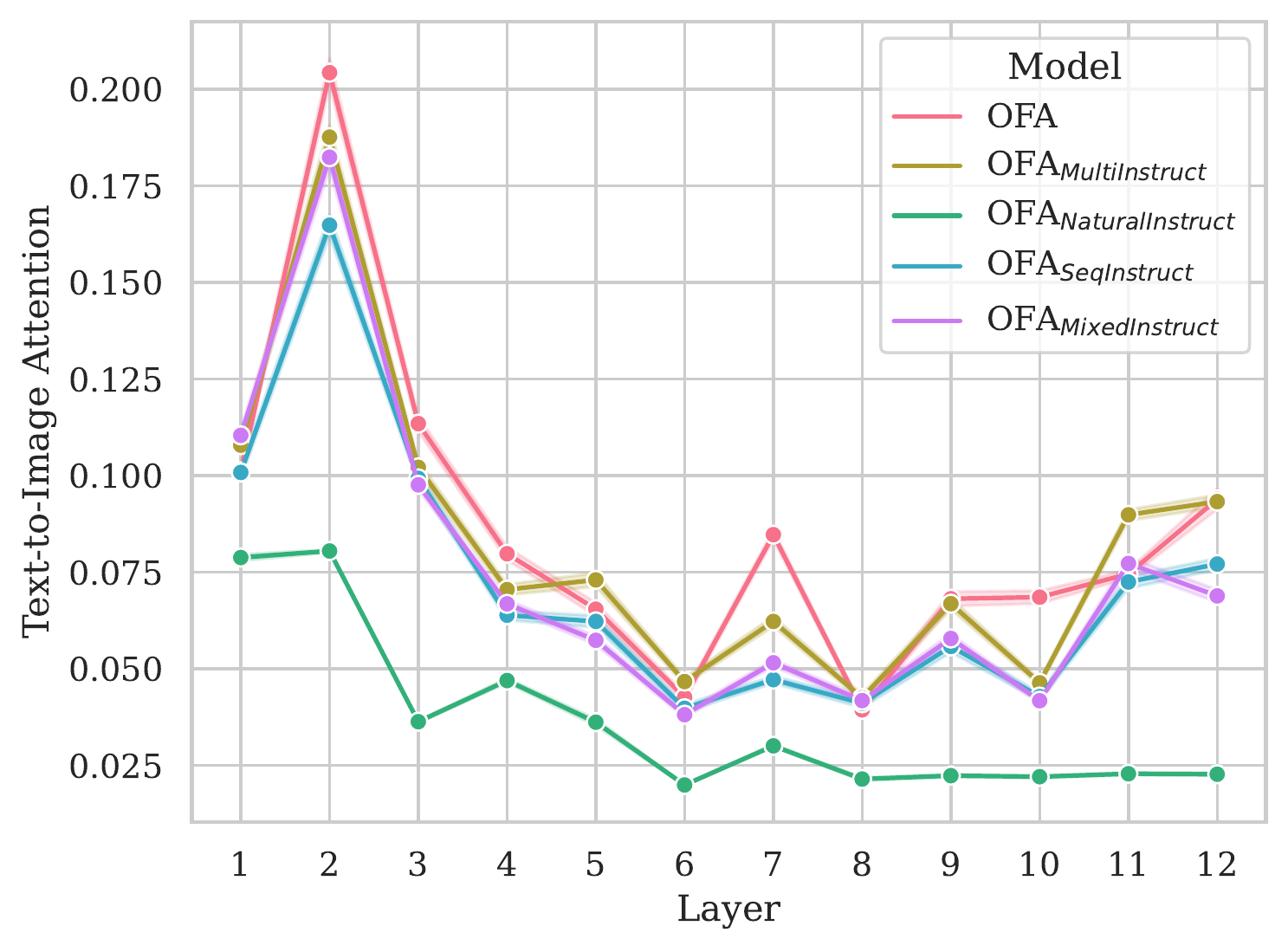}
        \caption{Visual Entailment}
    \end{subfigure}%
    \hfill
    \begin{subfigure}[b]{0.33\textwidth}
        \includegraphics[width=\textwidth]{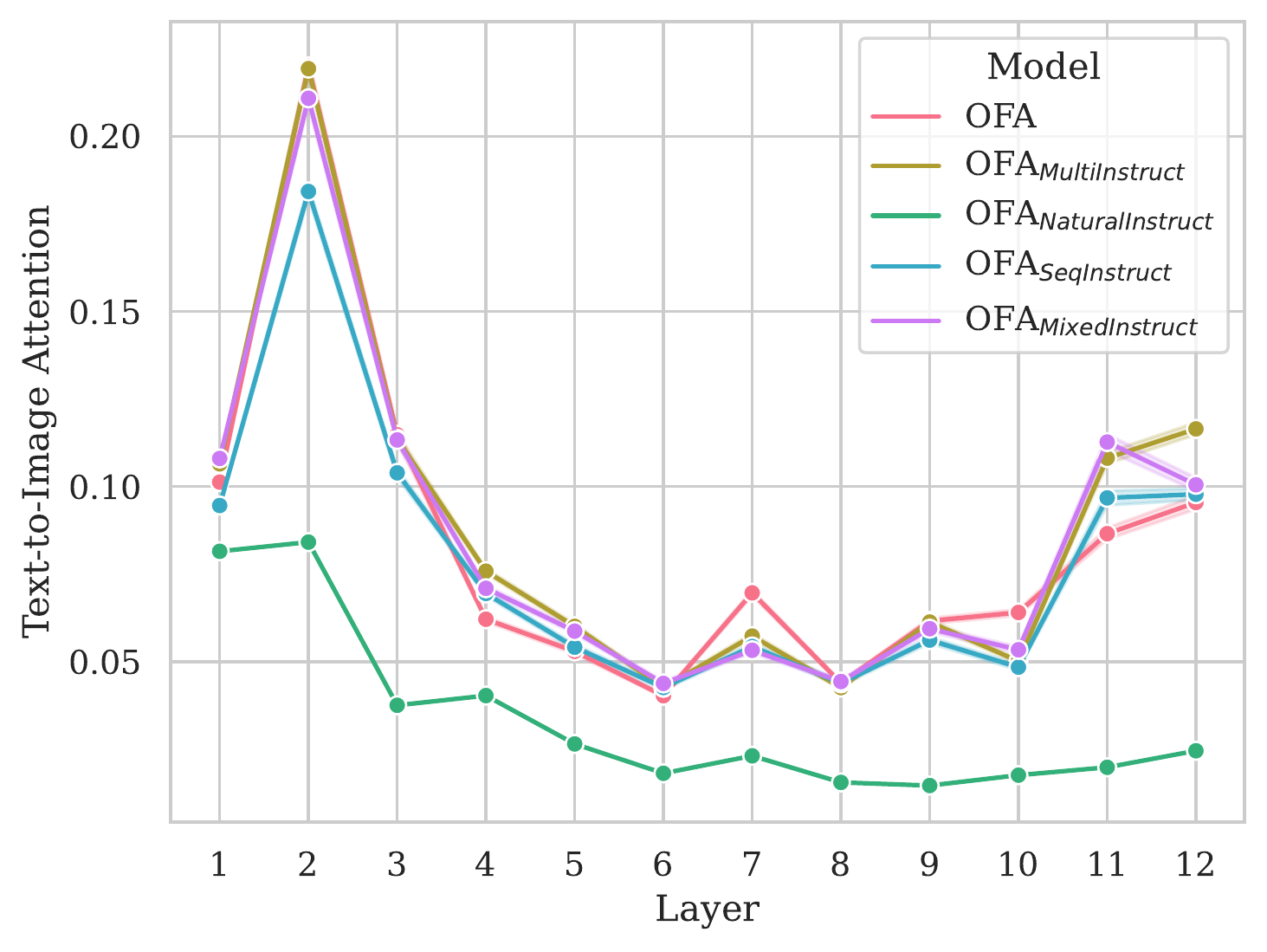}
        \caption{Visual Text Extraction}
    \end{subfigure}
    \caption{\textbf{Text-to-Image Attention of OFA Encoder.}}
    \label{fig:attn}
\end{figure*}

%% file: acl2023.bbl
\begin{thebibliography}{53}
\expandafter\ifx\csname natexlab\endcsname\relax\def\natexlab#1{#1}\fi

\bibitem[{Aghajanyan et~al.(2021)Aghajanyan, Gupta, Shrivastava, Chen,
  Zettlemoyer, and Gupta}]{Armen2021prefinetuning}
Armen Aghajanyan, Anchit Gupta, Akshat Shrivastava, Xilun Chen, Luke
  Zettlemoyer, and Sonal Gupta. 2021.
\newblock \href {https://doi.org/10.18653/v1/2021.emnlp-main.468} {Muppet:
  Massive multi-task representations with pre-finetuning}.
\newblock In \emph{Proceedings of the 2021 Conference on Empirical Methods in
  Natural Language Processing}, pages 5799--5811, Online and Punta Cana,
  Dominican Republic. Association for Computational Linguistics.

\bibitem[{Alam et~al.(2022)Alam, Alam, Hasan, Hasnat, Imran, Ofli
  et~al.}]{alam2022medic}
Firoj Alam, Tanvirul Alam, Md~Hasan, Abul Hasnat, Muhammad Imran, Ferda Ofli,
  et~al. 2022.
\newblock Medic: a multi-task learning dataset for disaster image
  classification.
\newblock \emph{Neural Computing and Applications}, pages 1--24.

\bibitem[{Alayrac et~al.(2022)Alayrac, Donahue, Luc, Miech, Barr, Hasson, Lenc,
  Mensch, Millican, Reynolds et~al.}]{alayrac2022flamingo}
Jean-Baptiste Alayrac, Jeff Donahue, Pauline Luc, Antoine Miech, Iain Barr,
  Yana Hasson, Karel Lenc, Arthur Mensch, Katie Millican, Malcolm Reynolds,
  et~al. 2022.
\newblock Flamingo: a visual language model for few-shot learning.
\newblock \emph{arXiv preprint arXiv:2204.14198}.

\bibitem[{Bao et~al.(2022)Bao, Dong, Piao, and Wei}]{bao2022beit}
Hangbo Bao, Li~Dong, Songhao Piao, and Furu Wei. 2022.
\newblock \href
  {https://www.microsoft.com/en-us/research/publication/beit-bert-pre-training-of-image-transformers/}
  {Beit: Bert pre-training of image transformers}.
\newblock In \emph{ICLR 2022}.

\bibitem[{Brown et~al.(2020)Brown, Mann, Ryder, Subbiah, Kaplan, Dhariwal,
  Neelakantan, Shyam, Sastry, Askell et~al.}]{brown2020language}
Tom Brown, Benjamin Mann, Nick Ryder, Melanie Subbiah, Jared~D Kaplan, Prafulla
  Dhariwal, Arvind Neelakantan, Pranav Shyam, Girish Sastry, Amanda Askell,
  et~al. 2020.
\newblock Language models are few-shot learners.
\newblock \emph{Advances in neural information processing systems},
  33:1877--1901.

\bibitem[{Chiu et~al.(2020)Chiu, Zhao, and Gurari}]{chiu2020assessing}
Tai-Yin Chiu, Yinan Zhao, and Danna Gurari. 2020.
\newblock Assessing image quality issues for real-world problems.
\newblock In \emph{Proceedings of the IEEE/CVF Conference on Computer Vision
  and Pattern Recognition}, pages 3646--3656.

\bibitem[{Cho et~al.(2021)Cho, Lei, Tan, and Bansal}]{cho2021unifying}
Jaemin Cho, Jie Lei, Hao Tan, and Mohit Bansal. 2021.
\newblock Unifying vision-and-language tasks via text generation.
\newblock In \emph{International Conference on Machine Learning}, pages
  1931--1942. PMLR.

\bibitem[{Das et~al.(2017)Das, Kottur, Gupta, Singh, Yadav, Moura, Parikh, and
  Batra}]{das2017visual}
Abhishek Das, Satwik Kottur, Khushi Gupta, Avi Singh, Deshraj Yadav,
  Jos{\'e}~MF Moura, Devi Parikh, and Dhruv Batra. 2017.
\newblock Visual dialog.
\newblock In \emph{Proceedings of the IEEE conference on computer vision and
  pattern recognition}, pages 326--335.

\bibitem[{Esser et~al.(2021)Esser, Rombach, and Ommer}]{esser2021taming}
Patrick Esser, Robin Rombach, and Bjorn Ommer. 2021.
\newblock Taming transformers for high-resolution image synthesis.
\newblock In \emph{Proceedings of the IEEE/CVF conference on computer vision
  and pattern recognition}, pages 12873--12883.

\bibitem[{Gemmeke et~al.(2017)Gemmeke, Ellis, Freedman, Jansen, Lawrence,
  Moore, Plakal, and Ritter}]{Gemmeke2017Audioset}
Jort~F. Gemmeke, Daniel P.~W. Ellis, Dylan Freedman, Aren Jansen, Wade
  Lawrence, R.~Channing Moore, Manoj Plakal, and Marvin Ritter. 2017.
\newblock \href {https://doi.org/10.1109/ICASSP.2017.7952261} {Audio set: An
  ontology and human-labeled dataset for audio events}.
\newblock In \emph{2017 {IEEE} International Conference on Acoustics, Speech
  and Signal Processing, {ICASSP} 2017, New Orleans, LA, USA, March 5-9, 2017},
  pages 776--780. {IEEE}.

\bibitem[{Goyal et~al.(2017)Goyal, Khot, Summers-Stay, Batra, and
  Parikh}]{goyal2017making}
Yash Goyal, Tejas Khot, Douglas Summers-Stay, Dhruv Batra, and Devi Parikh.
  2017.
\newblock Making the v in vqa matter: Elevating the role of image understanding
  in visual question answering.
\newblock In \emph{Proceedings of the IEEE conference on computer vision and
  pattern recognition}, pages 6904--6913.

\bibitem[{Gupta et~al.(2022)Gupta, Jiao, Yeh, Mehri, Eskenazi, and
  Bigham}]{prakhar2022instructdial}
Prakhar Gupta, Cathy Jiao, Yi-Ting Yeh, Shikib Mehri, Maxine Eskenazi, and
  Jeffrey~P. Bigham. 2022.
\newblock \href {https://doi.org/10.48550/ARXIV.2205.12673} {Improving zero and
  few-shot generalization in dialogue through instruction tuning}.

\bibitem[{Han et~al.(2022)Han, Zhao, Ding, Liu, and Sun}]{han2022ptr}
Xu~Han, Weilin Zhao, Ning Ding, Zhiyuan Liu, and Maosong Sun. 2022.
\newblock Ptr: Prompt tuning with rules for text classification.
\newblock \emph{AI Open}.

\bibitem[{Hudson and Manning(2019)}]{hudson2019gqa}
Drew~A Hudson and Christopher~D Manning. 2019.
\newblock Gqa: A new dataset for real-world visual reasoning and compositional
  question answering.
\newblock In \emph{Proceedings of the IEEE/CVF conference on computer vision
  and pattern recognition}, pages 6700--6709.

\bibitem[{Ionescu et~al.(2014)Ionescu, Papava, Olaru, and
  Sminchisescu}]{Ionescu2014video}
Catalin Ionescu, Dragos Papava, Vlad Olaru, and Cristian Sminchisescu. 2014.
\newblock \href {https://doi.org/10.1109/TPAMI.2013.248} {Human3.6m: Large
  scale datasets and predictive methods for 3d human sensing in natural
  environments}.
\newblock \emph{{IEEE} Trans. Pattern Anal. Mach. Intell.}, 36(7):1325--1339.

\bibitem[{Kafle and Kanan(2017)}]{kafle2017analysis}
Kushal Kafle and Christopher Kanan. 2017.
\newblock An analysis of visual question answering algorithms.
\newblock In \emph{Proceedings of the IEEE international conference on computer
  vision}, pages 1965--1973.

\bibitem[{Kiela et~al.(2020)Kiela, Firooz, Mohan, Goswami, Singh, Ringshia, and
  Testuggine}]{kiela2020hateful}
Douwe Kiela, Hamed Firooz, Aravind Mohan, Vedanuj Goswami, Amanpreet Singh,
  Pratik Ringshia, and Davide Testuggine. 2020.
\newblock The hateful memes challenge: Detecting hate speech in multimodal
  memes.
\newblock \emph{Advances in Neural Information Processing Systems},
  33:2611--2624.

\bibitem[{Krishna et~al.(2017)Krishna, Zhu, Groth, Johnson, Hata, Kravitz,
  Chen, Kalantidis, Li, Shamma et~al.}]{krishna2017visual}
Ranjay Krishna, Yuke Zhu, Oliver Groth, Justin Johnson, Kenji Hata, Joshua
  Kravitz, Stephanie Chen, Yannis Kalantidis, Li-Jia Li, David~A Shamma, et~al.
  2017.
\newblock Visual genome: Connecting language and vision using crowdsourced
  dense image annotations.
\newblock \emph{International journal of computer vision}, 123(1):32--73.

\bibitem[{Li et~al.(2022{\natexlab{a}})Li, Zhu, Jiang, Zhu, Li, Yuan, Wang,
  Qiao, Wang, Wang, and Dai}]{li2022UniPerceiverV2}
Hao Li, Jinguo Zhu, Xiaohu Jiang, Xizhou Zhu, Hongsheng Li, Chun Yuan, Xiaohua
  Wang, Yu~Qiao, Xiaogang Wang, Wenhai Wang, and Jifeng Dai.
  2022{\natexlab{a}}.
\newblock \href {https://doi.org/10.48550/arXiv.2211.09808} {Uni-perceiver v2:
  {A} generalist model for large-scale vision and vision-language tasks}.
\newblock \emph{CoRR}, abs/2211.09808.

\bibitem[{Li et~al.(2022{\natexlab{b}})Li, Gan, Lin, Lin, Liu, Liu, and
  Wang}]{li2022lavender}
Linjie Li, Zhe Gan, Kevin Lin, Chung-Ching Lin, Zicheng Liu, Ce~Liu, and Lijuan
  Wang. 2022{\natexlab{b}}.
\newblock Lavender: Unifying video-language understanding as masked language
  modeling.
\newblock \emph{arXiv preprint arXiv:2206.07160}.

\bibitem[{Li and Liang(2021)}]{Li2021prefixtuning}
Xiang~Lisa Li and Percy Liang. 2021.
\newblock \href {https://doi.org/10.18653/v1/2021.acl-long.353} {Prefix-tuning:
  Optimizing continuous prompts for generation}.
\newblock In \emph{Proceedings of the 59th Annual Meeting of the Association
  for Computational Linguistics and the 11th International Joint Conference on
  Natural Language Processing, {ACL/IJCNLP} 2021, (Volume 1: Long Papers),
  Virtual Event, August 1-6, 2021}, pages 4582--4597. Association for
  Computational Linguistics.

\bibitem[{Lin(2004)}]{lin2004rouge}
Chin-Yew Lin. 2004.
\newblock Rouge: A package for automatic evaluation of summaries.
\newblock In \emph{Text summarization branches out}, pages 74--81.

\bibitem[{Lin et~al.(2014)Lin, Maire, Belongie, Hays, Perona, Ramanan,
  Doll{\'a}r, and Zitnick}]{lin2014microsoft}
Tsung-Yi Lin, Michael Maire, Serge Belongie, James Hays, Pietro Perona, Deva
  Ramanan, Piotr Doll{\'a}r, and C~Lawrence Zitnick. 2014.
\newblock Microsoft coco: Common objects in context.
\newblock In \emph{European conference on computer vision}, pages 740--755.
  Springer.

\bibitem[{Liu et~al.(2022{\natexlab{a}})Liu, Emerson, and
  Collier}]{liu2022visual}
Fangyu Liu, Guy Emerson, and Nigel Collier. 2022{\natexlab{a}}.
\newblock Visual spatial reasoning.
\newblock \emph{arXiv preprint arXiv:2205.00363}.

\bibitem[{Liu et~al.(2022{\natexlab{b}})Liu, Tam, Muqeeth, Mohta, Huang,
  Bansal, and Raffel}]{Liu2022fewshot}
Haokun Liu, Derek Tam, Mohammed Muqeeth, Jay Mohta, Tenghao Huang, Mohit
  Bansal, and Colin Raffel. 2022{\natexlab{b}}.
\newblock \href {https://doi.org/10.48550/arXiv.2205.05638} {Few-shot
  parameter-efficient fine-tuning is better and cheaper than in-context
  learning}.
\newblock \emph{CoRR}, abs/2205.05638.

\bibitem[{Liu et~al.(2021)Liu, Yuan, Fu, Jiang, Hayashi, and
  Neubig}]{liu2021pre}
Pengfei Liu, Weizhe Yuan, Jinlan Fu, Zhengbao Jiang, Hiroaki Hayashi, and
  Graham Neubig. 2021.
\newblock Pre-train, prompt, and predict: A systematic survey of prompting
  methods in natural language processing.
\newblock \emph{arXiv preprint arXiv:2107.13586}.

\bibitem[{Lu et~al.(2022)Lu, Clark, Zellers, Mottaghi, and
  Kembhavi}]{lu2022unifiedIO}
Jiasen Lu, Christopher Clark, Rowan Zellers, Roozbeh Mottaghi, and Aniruddha
  Kembhavi. 2022.
\newblock \href {https://doi.org/10.48550/ARXIV.2206.08916} {Unified-io: A
  unified model for vision, language, and multi-modal tasks}.

\bibitem[{Marino et~al.(2019)Marino, Rastegari, Farhadi, and
  Mottaghi}]{marino2019ok}
Kenneth Marino, Mohammad Rastegari, Ali Farhadi, and Roozbeh Mottaghi. 2019.
\newblock Ok-vqa: A visual question answering benchmark requiring external
  knowledge.
\newblock In \emph{Proceedings of the IEEE/cvf conference on computer vision
  and pattern recognition}, pages 3195--3204.

\bibitem[{Min et~al.(2021)Min, Lewis, Zettlemoyer, and
  Hajishirzi}]{min2021metaicl}
Sewon Min, Mike Lewis, Luke Zettlemoyer, and Hannaneh Hajishirzi. 2021.
\newblock Metaicl: Learning to learn in context.
\newblock \emph{arXiv preprint arXiv:2110.15943}.

\bibitem[{Mishra et~al.(2022)Mishra, Khashabi, Baral, and
  Hajishirzi}]{Swaroop2022naturalinstructions}
Swaroop Mishra, Daniel Khashabi, Chitta Baral, and Hannaneh Hajishirzi. 2022.
\newblock \href {https://doi.org/10.18653/v1/2022.acl-long.244} {Cross-task
  generalization via natural language crowdsourcing instructions}.
\newblock In \emph{Proceedings of the 60th Annual Meeting of the Association
  for Computational Linguistics (Volume 1: Long Papers)}, pages 3470--3487,
  Dublin, Ireland. Association for Computational Linguistics.

\bibitem[{Panayotov et~al.(2015)Panayotov, Chen, Povey, and
  Khudanpur}]{Panayotov2015Librispeech}
Vassil Panayotov, Guoguo Chen, Daniel Povey, and Sanjeev Khudanpur. 2015.
\newblock \href {https://doi.org/10.1109/ICASSP.2015.7178964} {Librispeech: An
  {ASR} corpus based on public domain audio books}.
\newblock In \emph{2015 {IEEE} International Conference on Acoustics, Speech
  and Signal Processing, {ICASSP} 2015, South Brisbane, Queensland, Australia,
  April 19-24, 2015}, pages 5206--5210. {IEEE}.

\bibitem[{Pham et~al.(2021)Pham, Kafle, Lin, Ding, Cohen, Tran, and
  Shrivastava}]{pham2021learning}
Khoi Pham, Kushal Kafle, Zhe Lin, Zhihong Ding, Scott Cohen, Quan Tran, and
  Abhinav Shrivastava. 2021.
\newblock Learning to predict visual attributes in the wild.
\newblock In \emph{Proceedings of the IEEE/CVF Conference on Computer Vision
  and Pattern Recognition}, pages 13018--13028.

\bibitem[{Sanh et~al.(2022)Sanh, Webson, Raffel, Bach, Sutawika, Alyafeai,
  Chaffin, Stiegler, Raja, Dey, Bari, Xu, Thakker, Sharma, Szczechla, Kim,
  Chhablani, Nayak, Datta, Chang, Jiang, Wang, Manica, Shen, Yong, Pandey,
  Bawden, Wang, Neeraj, Rozen, Sharma, Santilli, F{\'{e}}vry, Fries, Teehan,
  Scao, Biderman, Gao, Wolf, and Rush}]{Sanh2022T0}
Victor Sanh, Albert Webson, Colin Raffel, Stephen~H. Bach, Lintang Sutawika,
  Zaid Alyafeai, Antoine Chaffin, Arnaud Stiegler, Arun Raja, Manan Dey,
  M~Saiful Bari, Canwen Xu, Urmish Thakker, Shanya~Sharma Sharma, Eliza
  Szczechla, Taewoon Kim, Gunjan Chhablani, Nihal~V. Nayak, Debajyoti Datta,
  Jonathan Chang, Mike~Tian{-}Jian Jiang, Han Wang, Matteo Manica, Sheng Shen,
  Zheng~Xin Yong, Harshit Pandey, Rachel Bawden, Thomas Wang, Trishala Neeraj,
  Jos Rozen, Abheesht Sharma, Andrea Santilli, Thibault F{\'{e}}vry, Jason~Alan
  Fries, Ryan Teehan, Teven~Le Scao, Stella Biderman, Leo Gao, Thomas Wolf, and
  Alexander~M. Rush. 2022.
\newblock \href {https://openreview.net/forum?id=9Vrb9D0WI4} {Multitask
  prompted training enables zero-shot task generalization}.
\newblock In \emph{The Tenth International Conference on Learning
  Representations, {ICLR} 2022, Virtual Event, April 25-29, 2022}.
  OpenReview.net.

\bibitem[{Sennrich et~al.(2016)Sennrich, Haddow, and
  Birch}]{sennrich2016neural}
Rico Sennrich, Barry Haddow, and Alexandra Birch. 2016.
\newblock Neural machine translation of rare words with subword units.
\newblock In \emph{Proceedings of the 54th Annual Meeting of the Association
  for Computational Linguistics (Volume 1: Long Papers)}, pages 1715--1725.

\bibitem[{Singh et~al.(2022)Singh, Hu, Goswami, Couairon, Galuba, Rohrbach, and
  Kiela}]{singh2022flava}
Amanpreet Singh, Ronghang Hu, Vedanuj Goswami, Guillaume Couairon, Wojciech
  Galuba, Marcus Rohrbach, and Douwe Kiela. 2022.
\newblock Flava: A foundational language and vision alignment model.
\newblock In \emph{Proceedings of the IEEE/CVF Conference on Computer Vision
  and Pattern Recognition}, pages 15638--15650.

\bibitem[{Singh et~al.(2019)Singh, Natarajan, Shah, Jiang, Chen, Batra, Parikh,
  and Rohrbach}]{singh2019towards}
Amanpreet Singh, Vivek Natarajan, Meet Shah, Yu~Jiang, Xinlei Chen, Dhruv
  Batra, Devi Parikh, and Marcus Rohrbach. 2019.
\newblock Towards vqa models that can read.
\newblock In \emph{Proceedings of the IEEE/CVF conference on computer vision
  and pattern recognition}, pages 8317--8326.

\bibitem[{Soomro et~al.(2012)Soomro, Zamir, and Shah}]{Soomro2012video}
Khurram Soomro, Amir~Roshan Zamir, and Mubarak Shah. 2012.
\newblock \href {http://arxiv.org/abs/1212.0402} {{UCF101:} {A} dataset of 101
  human actions classes from videos in the wild}.
\newblock \emph{CoRR}, abs/1212.0402.

\bibitem[{Suhr et~al.(2017)Suhr, Lewis, Yeh, and Artzi}]{suhr2017corpus}
Alane Suhr, Mike Lewis, James Yeh, and Yoav Artzi. 2017.
\newblock A corpus of natural language for visual reasoning.
\newblock In \emph{Proceedings of the 55th Annual Meeting of the Association
  for Computational Linguistics (Volume 2: Short Papers)}, pages 217--223.

\bibitem[{Tan and Bansal(2019)}]{tan2019lxmert}
Hao Tan and Mohit Bansal. 2019.
\newblock Lxmert: Learning cross-modality encoder representations from
  transformers.
\newblock \emph{arXiv preprint arXiv:1908.07490}.

\bibitem[{Veit et~al.(2016)Veit, Matera, Neumann, Matas, and
  Belongie}]{veit2016coco}
Andreas Veit, Tomas Matera, Lukas Neumann, Jiri Matas, and Serge Belongie.
  2016.
\newblock Coco-text: Dataset and benchmark for text detection and recognition
  in natural images.
\newblock \emph{arXiv preprint arXiv:1601.07140}.

\bibitem[{Wang et~al.(2022{\natexlab{a}})Wang, Yang, Men, Lin, Bai, Li, Ma,
  Zhou, Zhou, and Yang}]{wang2022unifying}
Peng Wang, An~Yang, Rui Men, Junyang Lin, Shuai Bai, Zhikang Li, Jianxin Ma,
  Chang Zhou, Jingren Zhou, and Hongxia Yang. 2022{\natexlab{a}}.
\newblock Unifying architectures, tasks, and modalities through a simple
  sequence-to-sequence learning framework.
\newblock \emph{arXiv preprint arXiv:2202.03052}.

\bibitem[{Wang et~al.(2022{\natexlab{b}})Wang, Yu, and Huang}]{wang2022art}
Sijia Wang, Mo~Yu, and Lifu Huang. 2022{\natexlab{b}}.
\newblock The art of prompting: Event detection based on type specific prompts.
\newblock \emph{arXiv preprint arXiv:2204.07241}.

\bibitem[{Wang et~al.(2022{\natexlab{c}})Wang, Bao, Dong, Bjorck, Peng, Liu,
  Aggarwal, Mohammed, Singhal, Som, and Wei}]{wang2022beit3}
Wenhui Wang, Hangbo Bao, Li~Dong, Johan Bjorck, Zhiliang Peng, Qiang Liu, Kriti
  Aggarwal, Owais~Khan Mohammed, Saksham Singhal, Subhojit Som, and Furu Wei.
  2022{\natexlab{c}}.
\newblock \href {https://doi.org/10.48550/arXiv.2208.10442} {Image as a foreign
  language: Beit pretraining for all vision and vision-language tasks}.
\newblock \emph{CoRR}, abs/2208.10442.

\bibitem[{Wang et~al.(2022{\natexlab{d}})Wang, Mishra, Alipoormolabashi, Kordi,
  Mirzaei, Arunkumar, Ashok, Dhanasekaran, Naik, Stap, Pathak, Karamanolakis,
  Lai, Purohit, Mondal, Anderson, Kuznia, Doshi, Patel, Pal, Moradshahi,
  Parmar, Purohit, Varshney, Kaza, Verma, Puri, Karia, Sampat, Doshi, Mishra,
  Reddy, Patro, Dixit, Shen, Baral, Choi, Smith, Hajishirzi, and
  Khashabi}]{Wang2022superNaturalInstruct}
Yizhong Wang, Swaroop Mishra, Pegah Alipoormolabashi, Yeganeh Kordi, Amirreza
  Mirzaei, Anjana Arunkumar, Arjun Ashok, Arut~Selvan Dhanasekaran, Atharva
  Naik, David Stap, Eshaan Pathak, Giannis Karamanolakis, Haizhi~Gary Lai,
  Ishan Purohit, Ishani Mondal, Jacob Anderson, Kirby Kuznia, Krima Doshi,
  Maitreya Patel, Kuntal~Kumar Pal, Mehrad Moradshahi, Mihir Parmar, Mirali
  Purohit, Neeraj Varshney, Phani~Rohitha Kaza, Pulkit Verma, Ravsehaj~Singh
  Puri, Rushang Karia, Shailaja~Keyur Sampat, Savan Doshi, Siddhartha Mishra,
  Sujan Reddy, Sumanta Patro, Tanay Dixit, Xudong Shen, Chitta Baral, Yejin
  Choi, Noah~A. Smith, Hannaneh Hajishirzi, and Daniel Khashabi.
  2022{\natexlab{d}}.
\newblock \href {https://doi.org/10.48550/ARXIV.2204.07705}
  {Super-naturalinstructions: Generalization via declarative instructions on
  1600+ nlp tasks}.

\bibitem[{Webson and Pavlick(2022)}]{pavlick2022promptwording}
Albert Webson and Ellie Pavlick. 2022.
\newblock \href {https://doi.org/10.18653/v1/2022.naacl-main.167} {Do
  prompt-based models really understand the meaning of their prompts?}
\newblock In \emph{Proceedings of the 2022 Conference of the North American
  Chapter of the Association for Computational Linguistics: Human Language
  Technologies}, pages 2300--2344, Seattle, United States. Association for
  Computational Linguistics.

\bibitem[{Wei et~al.(2021)Wei, Bosma, Zhao, Guu, Yu, Lester, Du, Dai, and
  Le}]{wei2021flan}
Jason Wei, Maarten Bosma, Vincent~Y. Zhao, Kelvin Guu, Adams~Wei Yu, Brian
  Lester, Nan Du, Andrew~M. Dai, and Quoc~V. Le. 2021.
\newblock \href {http://arxiv.org/abs/2109.01652} {Finetuned language models
  are zero-shot learners}.
\newblock \emph{CoRR}, abs/2109.01652.

\bibitem[{Xie et~al.(2019)Xie, Lai, Doran, and Kadav}]{xie2019visual}
Ning Xie, Farley Lai, Derek Doran, and Asim Kadav. 2019.
\newblock Visual entailment: A novel task for fine-grained image understanding.
\newblock \emph{arXiv preprint arXiv:1901.06706}.

\bibitem[{Xie et~al.(2021)Xie, Raghunathan, Liang, and
  Ma}]{xie2021explain-in-context}
Sang~Michael Xie, Aditi Raghunathan, Percy Liang, and Tengyu Ma. 2021.
\newblock \href {http://arxiv.org/abs/2111.02080} {An explanation of in-context
  learning as implicit bayesian inference}.
\newblock \emph{CoRR}, abs/2111.02080.

\bibitem[{Yao et~al.(2022)Yao, Shah, Sun, Cho, and Huang}]{yao2022end}
Barry~Menglong Yao, Aditya Shah, Lichao Sun, Jin-Hee Cho, and Lifu Huang. 2022.
\newblock End-to-end multimodal fact-checking and explanation generation: A
  challenging dataset and models.
\newblock \emph{arXiv preprint arXiv:2205.12487}.

\bibitem[{You et~al.(2022)You, Chen, Liu, Ge, Wu, and Zou}]{You2022audioData}
Chenyu You, Nuo Chen, Fenglin Liu, Shen Ge, Xian Wu, and Yuexian Zou. 2022.
\newblock \href {https://doi.org/10.18653/v1/2022.findings-naacl.91}
  {End-to-end spoken conversational question answering: Task, dataset and
  model}.
\newblock In \emph{Findings of the Association for Computational Linguistics:
  {NAACL} 2022, Seattle, WA, United States, July 10-15, 2022}, pages
  1219--1232. Association for Computational Linguistics.

\bibitem[{Yu et~al.(2016)Yu, Poirson, Yang, Berg, and Berg}]{yu2016modeling}
Licheng Yu, Patrick Poirson, Shan Yang, Alexander~C Berg, and Tamara~L Berg.
  2016.
\newblock Modeling context in referring expressions.
\newblock In \emph{European Conference on Computer Vision}, pages 69--85.
  Springer.

\bibitem[{Zellers et~al.(2019)Zellers, Bisk, Farhadi, and
  Choi}]{zellers2019recognition}
Rowan Zellers, Yonatan Bisk, Ali Farhadi, and Yejin Choi. 2019.
\newblock From recognition to cognition: Visual commonsense reasoning.
\newblock In \emph{Proceedings of the IEEE/CVF conference on computer vision
  and pattern recognition}, pages 6720--6731.

\bibitem[{Zhu et~al.(2016)Zhu, Groth, Bernstein, and Fei-Fei}]{zhu2016visual7w}
Yuke Zhu, Oliver Groth, Michael Bernstein, and Li~Fei-Fei. 2016.
\newblock Visual7w: Grounded question answering in images.
\newblock In \emph{Proceedings of the IEEE conference on computer vision and
  pattern recognition}, pages 4995--5004.

\end{thebibliography}
